\documentclass[10pt,journal,compsoc]{IEEEtran}

\ifCLASSOPTIONcompsoc
  \usepackage[nocompress]{cite}
\else
  \usepackage{cite}
\fi
\ifCLASSINFOpdf
\else
\fi
\pdfoutput=1

\usepackage{url}
\usepackage{times}
\usepackage{epsfig}
\usepackage{graphicx}
\usepackage{amsmath}
\usepackage{amssymb}
\usepackage{multirow}
\usepackage{colortbl}
\usepackage{color}
\usepackage{threeparttable}
\usepackage{booktabs}
\usepackage{adjustbox}
\usepackage{caption}
\usepackage{subcaption}
\usepackage[misc]{ifsym}
\usepackage{xcolor,colortbl}
\usepackage{makecell}
\usepackage{stfloats}
\newcommand*{\belowrulesepcolor}[1]{%
  \noalign{%
    \kern-\belowrulesep 
    \begingroup 
      \color{#1}%
      \hrule height\belowrulesep 
    \endgroup 
    \vspace{-0.03mm}
  }%
} 
\newcommand*{\aboverulesepcolor}[1]{%
  \noalign{%
  \vspace{-0.03mm}
    \begingroup 
      \color{#1}%
      \hrule height\aboverulesep 
    \endgroup 
    \kern-\aboverulesep 
  }%
}

\usepackage{pifont}

\usepackage{xspace}
\makeatletter
\DeclareRobustCommand\onedot{\futurelet\@let@token\@onedot}
\def\@onedot{\ifx\@let@token.\else.\null\fi\xspace}

\usepackage[linesnumbered,lined,boxed,commentsnumbered,ruled]{algorithm2e}
\usepackage[pagebackref=true,breaklinks=true,colorlinks,citecolor=blue,urlcolor=blue,linkcolor=blue,bookmarks=false]{hyperref}
\begin{document}
%


\title{DC-SAM: In-Context Segment Anything in Images and Videos via Dual Consistency}

\author{
        Mengshi Qi,
        Pengfei~Zhu,
        Xiangtai Li,
        Xiaoyang Bi,
        Lu Qi,
        Huadong Ma,
        Ming-Hsuan Yang
\IEEEcompsocitemizethanks{
\IEEEcompsocthanksitem M. Qi, P. Zhu, X. Bi and H. Ma are with the State Key Laboratory of Networking and Switching Technology, Beijing University of Posts and Telecommunications, China.
\IEEEcompsocthanksitem X. Li is with Nanyang Technological University, Singapore. 
\IEEEcompsocthanksitem  Lu. Qi and M.-H. Yang are with the UC Merced, US. 
}
}


\IEEEtitleabstractindextext{
\begin{abstract}
Given a single labeled examples, in-context segmentation aims to segment corresponding objects. 
This setting, known as one-shot segmentation in few-shot learning, explores the segmentation model's generalization ability and has been applied to various vision tasks, including scene understanding and image/video editing.
While recent Segment Anything Models (SAMs) have achieved state-of-the-art results in interactive segmentation, these approaches are not directly applicable to in-context segmentation.
%
%
In this work, we propose the Dual Consistency SAM 
(DC-SAM) method based on prompt-tuning to adapt SAM and SAM2 for in-context segmentation of both images and videos. 
Our key insights are to enhance the features of the SAM's prompt encoder in segmentation by providing high-quality visual prompts.
When generating a mask prior from support images, we fuse the SAM features to better align the prompt encoder rather than relying solely on a pre-trained backbone.
Then, we design a cycle-consistent cross-attention on fused features and initial visual prompts.
This design leverages coarse masks from the SAM mask decoder to ensure consistency between features and visual prompts.
Next, a dual-branch design is provided by using the discriminative positive and negative prompts in the prompt encoder.
Furthermore, we design a simple mask-tube training strategy to adopt our proposed dual consistency method into the mask tube.
Although the proposed DC-SAM is primarily designed for images, it can be seamlessly extended to the video domain with the support of SAM2.
%
%
Given the absence of in-context segmentation in the video domain, we manually curate and construct the first benchmark from existing video segmentation datasets, named \emph{In-Context Video Object Segmentation~(IC-VOS)}, to better assess the in-context capability of the model.
%
%
Extensive experiments demonstrate that our method achieves 55.5 (+1.4) mIoU on COCO-20$^i$, 73.0 (+1.1) mIoU on PASCAL-5$^i$, and a $\mathcal{J\&F}$ score of 71.52 on the proposed IC-VOS benchmark.
Our source code and benchmark are available at \url{https://github.com/zaplm/DC-SAM}.
\end{abstract}

\begin{IEEEkeywords}
Segment Anything Model, In-context Segmentation, Prompt Generation, Efficient Parameter Tuning
\end{IEEEkeywords}}

\maketitle

\IEEEdisplaynontitleabstractindextext

%
\IEEEpeerreviewmaketitle

\begin{figure*}[!htbp]
   \begin{center}
      \includegraphics[width=1\linewidth]{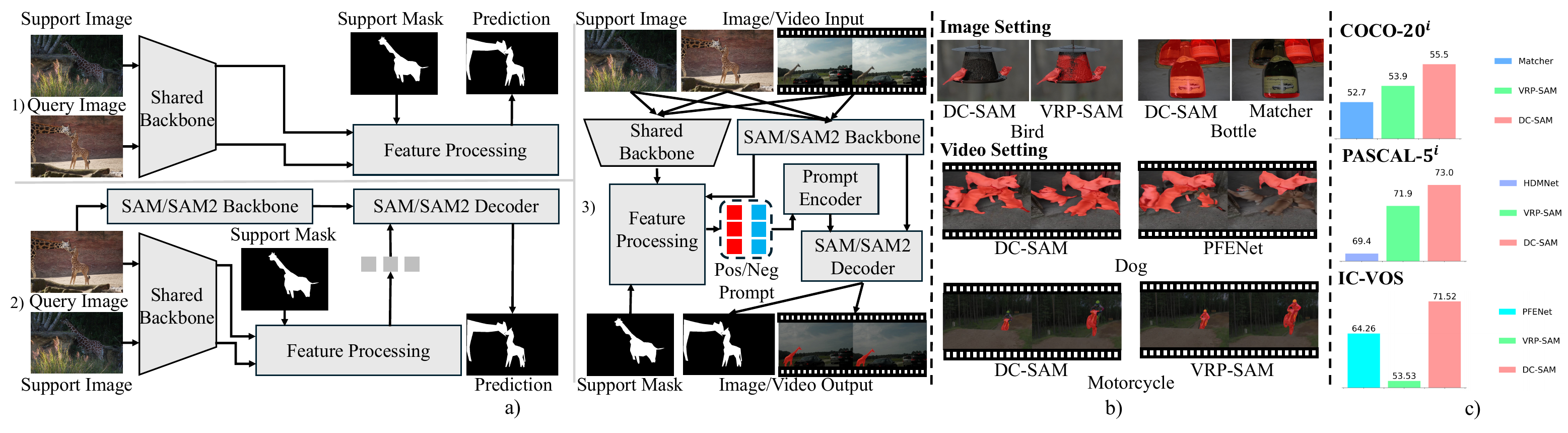}
   \end{center}
   \vspace{-4mm}
      \caption{Overview of the proposed DC-SAM method and IC-VOS benchmark. a) Comparison of the previous few-shot segmentation methods in 1), existing methods based on SAM/SAM2 in 2), and DC-SAM in 3). DC-SAM leverages multi-source features and generates positive and negative prompts by ensuring prompt consistency, integrating with SAM/SAM2 to achieve in-context segmentation for both images and videos; b) Visualization of image and video settings by DC-SAM; c) Quantitative comparison of DC-SAM with state-of-the-art approaches in terms of mIoU on COCO-20$^i$ and PASCAL-5$^i$, $\mathcal{J\&F}$ on the IC-VOS benchmark.}
      \vspace{-3mm}
   \label{fig:teaser}
\end{figure*}

\section{Introduction}
\label{sec:intro}

Recent visual foundation models such as Segment Anything models (SAM and SAM2)~\cite{kirillov2023segment,ravi2024sam2} have attracted significant attention in recent years.
Leveraging tens of millions of images and videos along with over a billion masks, the SAM series demonstrates strong performance in interactive segmentation and proves to be a valuable tool across a wide range of applications, including medical image segmentation~\cite{sam_medical}, open-vocabulary segmentation~\cite{ov_seg_sam,zhou2023rethinking}, and more.
In particular, numerous efforts~\cite{lisa,yuan2024ovsam,sa2va} have been made to exploit the versatile segmentation capabilities of SAMs in specific domains, such as reasoning~\cite{lisa}, extending the recognition ability, or combining with large language models~\cite{qi2024generalizable}.


Despite the state-of-the-art segmentation capabilities, SAMs lack the inherent ability to segment instances of the same category across multiple images given a single instance prompt, an ability we refer to as in-context segmentation inspired by the NLP domain~\cite{cot,ic_survey}.
In this task, the given image and object masks are called support or in-context examples, while the input images are called query images. 
Previous works explore such abilities with few-shot learning approaches~\cite{pfenet,bam,cyctr,hdmnet,amnet,abcb,DCAMA}, such as calculating the matching distance between query images and support images (known as visual prompts for in-context learning) or modeling object prototypes for better alignment.
%
However, these methods explore in-context learning ability with only a few examples and fail to generalize across diverse domains.
Most recently, several models~\cite{painter,seggpt,fang2024explore,wang2023skeleton} based on in-context learning have been developed where the prompts consist of input-output pairs of visual tasks.
Specifically, SegGPT~\cite{seggpt} explores in-context segmentation by co-training massive image-mask pairs, leading to good generalization ability across various one-shot segmentation benchmarks.
Although this work achieves promising performance, substantial computational resources and extensive annotated segmentation data are required to build such a system.
%
%
In contrast, prompt tuning~\cite{persam,matcher,vrpsam} offers an effectively alternative adaptive approach that has shown impressive generalization capabilities across diverse domains. 
%
%
This work proposes an adaptive SAM model for in-context image and video segmentation based on prompt tuning. 



We note that the existing approaches developed based on SAM~\cite{matcher,vrpsam} mainly depend on the characteristics extracted from the backbone networks and do not fully take advantage of the distinctive properties of the SAM-derived representations, as shown in Figure~\ref{fig:teaser}(a).
This limitation leads to an oversight of the differences between SAM and backbone network features during the prompt generation process, significantly impacting the accuracy of the generated results. 
Solely relying on SAM-extracted features for prompt generation often lacks sufficient utilization of the semantic priors of target categories, another critical factor contributing to suboptimal performance. 
%
%
In addition, no suitable benchmarks currently exist to evaluate the in-context segmentation capabilities of video data.
Existing benchmarks~\cite{youtubevos,davis,lvos,lvosv2,mose} for video segmentation focus mainly on segmenting and tracking pixels over time.
To the best of our knowledge, there are no benchmarks for evaluating this ability using in-context examples. 


To address the aforementioned problems, we construct the first \emph{In-Context Video Object Segmentation~(IC-VOS)} benchmark.
We collect examples from existing video segmentation benchmarks and in-context examples from COCO dataset, visually inspecting each example. 
The IC-VOS dataset comprises 369 videos, averaging 270 frames per video, totaling 99,549 frames across 30 semantic classes.
We then benchmark representative methods with SAM2 and propagate the masks from SAM2 to establish the first in-context video benchmarks.

For model design, we propose a novel feature extraction strategy for the prompt generation of SAM, termed~\emph{prompt consistency generation}.
The meta-architecture is shown in Figure~\ref{fig:teaser}(b).
Our approach first fuses features extracted by the SAM encoder with those obtained from the backbone network to generate a prior mask for the query image.   
Experimental results show that combining these two types of features significantly improves model performance.
In addition, we design two improvements involving positive and negative features of the SAM's prompt encoder. 
First, we employ a dual-branch strategy that generates positive and negative prompts using foreground and background masks, respectively. 
The main idea is to utilize the positive and negative prompt embeddings within SAM's prompt encoder to assign labels to the automatically generated prompts. 
By leveraging the interaction between positive and negative prompts, we achieve fine-grained control over the generated masks, since more confidential visual cues are provided.
Second, we incorporate a Cyclic Consistent Cross-Attention mechanism into the prompt generation process. 
This mechanism ensures semantic label consistency between input features and queries by aligning highly relevant support pixels with their corresponding categories. 
It effectively suppresses the propagation of conflicting semantic information. 
Consequently, this approach ensures that the generated prompts accurately focus on the most critical regions.
The proposed prompt consistency method can be easily extended to the video domain, particularly for the SAM2 architecture.
Specifically, we design a simple mask tube supervision by extending prompt consistency to the tube mask format.
Since our work can be applied to both SAM and SAM2, we term our method DC-SAM, an extension of the SAM series for in-context segmentation of images and videos.

The main contributions of this work are: 
\begin{itemize}
\item We propose a novel prompt-consistency method based on SAM, called Dual-Consistency SAM (DC-SAM), tailored for one-shot segmentation tasks. 
It exploits the positive and negative features of the visual prompts, leading to high-quality prompts for in-context segmentation.
Furthermore, this design can be easily extended to video tasks by combining the SAM and a new mask tube design.

\item We introduce a novel cyclic consistent cross-attention mechanism that ensures the final generated prompts better focus on the key regions requiring prompting. 
When combined with SAM, this mechanism effectively filters out potentially ambiguous components in the features, further enhancing the accuracy and specificity of in-context segmentation.

\item We collect a new video in-context segmentation benchmark, IC-VOS (In-Context Video Object Segmentation), featuring manually curated examples sourced from existing video benchmarks. 
In addition, we benchmark several representative works in {IC-VOS}.


\item With extensive experiments and ablation studies, the proposed method achieves state-of-the-art performance on various datasets and our newly proposed in-context segmentation benchmarks. DC-SAM achieves 55.5 (+1.4) mIoU on COCO-20$^i$, 73.0 (+1.1) mIoU on PASCAL-5$^i$, and a $\mathcal{J\&F}$ score of 71.52 on the IC-VOS benchmark.

\end{itemize}

\section{Related Work}
\label{sec:related_work}

\noindent
\textbf{Segmentation Anything Model.} SAM models~\cite{kirillov2023segment, ravi2024sam2} are proposed to segment objects in both image and video interactively. 
Using large-scale data co-training, SAM~\cite{kirillov2023segment} presents a novel data engine and a portable model for segmentation. 
Subsequent research has utilized SAM as an interactive segmentation tool for various vision tasks, including visual grounding~\cite{liu2023grounding}, tracking~\cite{cheng2023segmenttracking}, distillation and efficiency modeling~\cite{zhou2023edgesam}, medical analysis~\cite{wu2023medical,chen2023ma}, scene understanding~\cite{lv_neurips,qi_tip,STC-GAN,qi_CVPR19,qi_tcsvt} and image generation~\cite{persam}. 
%
%
To adapt SAM for few-shot and in-context segmentation, PerSAM~\cite{persam} and Matcher~\cite{matcher} employ patch cosine similarity to identify similar regions for subsequent tasks.
VRP-SAM~\cite{vrpsam} employs a comparable feature extraction method to identify similar regions for few-shot segmentation. 
It proposes a query-based approach based on cross-attention to generate prompts. 
In contrast, our method emphasizes the prompt encoding process in SAM, treating both foreground and background masks as essential constraints.
By leveraging positive and negative prompt branches along with a Cyclic Consistent Cross-Attention mechanism, DC-SAM efficiently utilizes SAM's prompt encoder and mask decoder to improve its in-context segmentation performance.

\noindent
\textbf{Few-shot Segmentation.} 
The goal of few-shot segmentation~\cite{pfenet} is to segment novel semantic categories by giving only a few annotated examples, including few-shot instance segmentation~\cite{fan2020fgn}, incremental few-shot instance segmentation~\cite{ganea2021incremental,han2024reference}, and generalized few-shot segmentation~\cite{li2021towards,han2024reference}. 
Our method primarily focuses on one-shot semantic segmentation, also known as in-context segmentation. This task aims to segment query images using only a single support sample.
%
%
Recent methods~\cite{pfenet,liu2022intermediate,liu2022learning,liu2022dynamic,li2021adaptive} are typically developed based on metric learning by matching spatial location features with semantic centroids. 
In particular, PFENet~\cite{pfenet} employs the last-layer features to generate prior masks and utilizes the mid-level features for segmentation.
CycTR~\cite{cyctr} introduces cyclic consistency between query and support features within the same class mask regions.
%
%
Then, two-branch conditional networks~\cite{shaban2017one,Rakelly2018ConditionalNF,lu2021simpler}, 4D dense convolution networks~\cite{vat,min2021hypercorrelation}, data augmentation methods~\cite{tritrong2021repurposing,zhang2021datasetgan}.
Recently, transformer-based models~\cite{zhang2022feature,shi2022dense,xie2021few,kim2023universal} have also been applied to solve the problem.
The core idea behind these works is to learn the correspondence in the feature space.
In addition, several few-shot segmentation methods are developed based on in-context mask generation~\cite{bar2022visual, painter, seggpt} and latent diffusion models~\cite{aligndiff}. 
However, existing approaches focus on directly designing a model to complete the entire few-shot segmentation task, relying on complex computational frameworks (\emph{e.g.}, Transformer or diffusion models). 
As a result, the generalizability of these models is limited.
Moreover, these methods do not fully utilize the general segmentation prior knowledge inherent in pre-trained foundational models such as SAM~\cite{kirillov2023segment}.
%
By integrating rich knowledge extracted from SAM and pre-trained backbones (such as ResNet~\cite{he2016deep} or the vision transformer-based DINO-v2~\cite{oquab2023dinov2}), our approach generalizes well across various benchmarks while incurring lower computational costs.




%
%


\noindent
\textbf{Prompt Tuning in Vision.} Due to the limited computation resources, most current work cannot re-train the foundation models from scratch. 
Inspired by prompt tuning in NLP, several methods~\cite{houlsby2019parameter,li2021prefix,hu2021lora,zaken2021bitfit,guo2020parameter,lester2021power,wu2024lgvi,yuan2024ovsam} fine-tune only a tiny portion of parameters to adapt the pre-trained foundation models to various downstream tasks, including image classification, segmentation, image generation and editing. 
The primary objective of these methods is to optimize the prompting process using the inherent knowledge in the model and then use the refined prompts to enhance performance. 
This approach preserves the original capabilities of the model and improves its effectiveness in specific tasks.
Our work belongs to the prompt tuning approach on SAM, where only the parameters in the prompt encoder are learned. 
In particular, we aim to learn better correspondence between objects in query and support images, facilitating the model to achieve strong performance in in-context segmentation. 

\noindent
\textbf{Video Segmentation Benchmarks.}  Large-scale video object segmentation (VOS) datasets, such as DAVIS~\cite{davis}, MOSE~\cite{mose}, and LVOS~\cite{lvos}, are widely used for numerous tasks.  
In these datasets, the mask of the target object is provided in the first frame, and subsequent frames require the segmentation of the corresponding objects throughout the video. 
However, in practical applications, when a specific semantic mask (provided along with a support image) needs to be segmented throughout the entire video, existing methods cannot inherently perform this task without requiring annotations in the first frame. 
%
Even with the assistance of recent models such as SAM~\cite{kirillov2023segment}, manual clicks or selections of the target object and semantics still require corrections to the edges or interiors, which is cumbersome and labor-intensive. 
To address these limitations, in this work, we collect the first in-context VOS benchmark to utilize a small number of images as prompts to achieve video semantic segmentation.

\begin{figure*}[h]
   \begin{center}
      \includegraphics[width=1.0\textwidth]{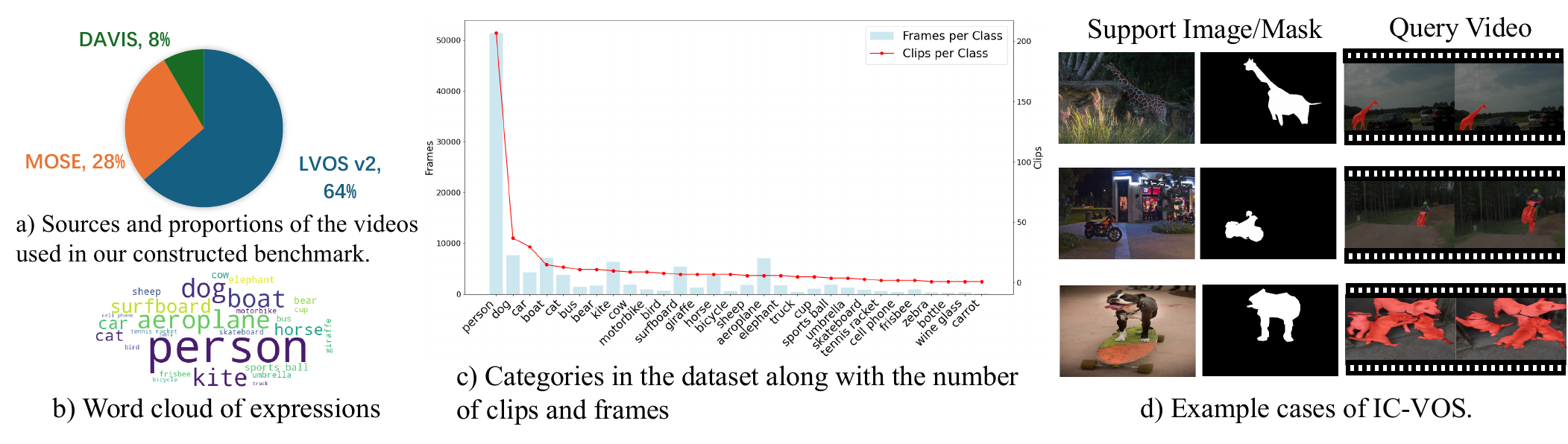}
   \end{center}
   \vspace{-5mm}
      \caption{Overview of our constructed IC-VOS benchmark. a) Distribution of video sources and their proportions. b) Word cloud of expressions. c) Categories in the dataset with the number of clips and frames for each category. d) Example cases illustrating the support image, support mask, and query video.}
   \label{fig:dataset_source}
   \vspace{-2mm}
\end{figure*}

\section{In-Context VOS Benchmark}
\label{sec:problem_vos_benchmarks}

%
%

We propose a new in-context video segmentation dataset. 
The goal is to enable segmentation models to automatically identify and segment target semantics in videos, eliminating the need for manual annotation of the first frame.

\noindent
\subsection{Problem Setting}

We first introduce the previous image setting.
The \emph{in-context image segmentation} task takes query images, support images, and support masks as inputs.
Our model is trained on a dataset $D_{\text{train}}$ and evaluated on $D_{\text{test}}$, where $D_{\text{train}}$ and $D_{\text{test}}$ contain mutually exclusive categories $C_{\text{train}}$ and $C_{\text{test}}$, respectively (\textit{i.e.}, $C_{\text{train}} \cap C_{\text{test}} = \emptyset$). 
Similar to prior work~\cite{pfenet,hdmnet,cyctr}, the original datasets are partitioned into multiple folds, each of which comprises a training set $D_{\text{train}}$ and a test set $D_{\text{test}}$. 
For each fold, there exists a query set $Q = \{(I_i^q, M_i^q)\}_{i=1}^{c}$ and a support set $S = \{(I_i^s, M_i^s)\}_{i=1}^{c}$, where $c$ denotes the total number of categories in the training set $D_{\text{train}}$ or the test set $D_{\text{test}}$. 
Here, $I$ and $M$ represent RGB images and their corresponding segmentation masks.
The objective is to produce a mask output $\hat{M} \in \mathbb{R}^{H \times W \times 1}$, given a query image $I^q \in \mathbb{R}^{H \times W \times 3}$ and a support set $S$, where $H$ and $W$ denote the height and width dimensions.


For \emph{in-context video segmentation}, a given image and its corresponding semantic mask are used to segment the associated semantics within a video clip.
The image and its mask play the same role as the support image and support mask in in-context image segmentation. 
%
%
Given a support image $I^s \in\mathbb{R}^{H \times W \times 3}$ and its corresponding mask $M^s \in \mathbb{R}^{H \times W \times 1}$, the task involves segmenting each frame of the query video $V^q \in \mathbb{R}^{T \times H \times W \times 3}$ to obtain the associated semantic masks, resulting in a mask tube $M_{pred}^q \in \mathbb{R}^{T \times H \times W \times 1}$, where $T$ means the frame numbers of the video.
Thus, this task requires the model to segment and track objects with the same semantics as the given support object, posing greater challenges compared to in-context image segmentation.

\begin{table}[t]
    \centering
        \small
        \captionof{table}{Comparison of the video portion of our proposed benchmark with other well-known VOS datasets. Annotation type indicates whether a mask \textbf{M} or a bounding box \textbf{B} is provided.}
        \vspace{-1mm}
        \resizebox{0.49\textwidth}{!}{
        \begin{tabular}{c|ccccc} 
            \toprule
            \multirow{2}{*}{Dataset} & \multirow{2}{*}{Videos} & Mean & Total & \multirow{2}{*}{Classes} & Annotations \\
            &&Frames&Frames&&Type \\ \midrule
            DAVIS~\cite{davis} & 90 & 69 & 6298 & - & \textbf{M}  \\ 
            MOSE~\cite{mose} & 2149 & 73 & 159,600 & 36 & \textbf{M} \\ 
            LVOS~\cite{lvos} & 220 & 576 & 126,280 & 27 & \textbf{M} \\
            LVOS v2~\cite{lvosv2} & 720 & 412 & 296,401 & 44 & \textbf{M} \\ 
            YouTube-VOS\cite{youtubevos} & 4453 & 27 & 120,532& 94 & \textbf{M} \\
            UAV20L~\cite{uav20l} & 20 & 2934 & 59,000 & 5 & \textbf{B}  \\ \midrule
            IC-VOS (Ours) & 369 & 270 & 99,549 & 30 & \textbf{M} \\
            \bottomrule
        \end{tabular}}
        \label{tab:statistics}
        \vspace{-1mm}
\end{table}

\subsection{Dataset Construction}

The benchmark consists of image data and video datasets. 
The image dataset is divided into a training set and a validation set. 
%
%
%
%
%
%
We use the training set of the COCO semantic segmentation dataset~\cite{lin2014microsoft} as the training set for the proposed benchmark. 
The COCO dataset is renowned for its richness and diversity. 
It contains a vast number of images and detailed semantic annotations, which provide a solid foundation for training our model. 
Additionally, the COO validation set serves as the source of image prompts during our validation process. 

To reduce annotation costs and leverage high-quality annotations from existing VOS datasets, we use the DAVIS~\cite{davis}, MOSE~\cite{mose}, and LVOS v2~\cite{lvosv2} datasets and their annotations as a template to construct our benchmark. 
However, these VOS datasets use instance-level annotations, which differ from our requirements for semantic-level segmentation. 
As such, we manually screen the video data from these three datasets to meet the following rules. 
First, all instances of a given category that appear in the video are annotated in the VOS dataset, at least in the first frame. 
Second, the categories of the instances must belong to one of the 80 classes in the COCO instance segmentation dataset.

We overlay the annotations from all the videos and save them as GIFs to facilitate manual selection. 
Only videos satisfying the above criteria will be selected.
We archive the annotations for each instance into $n$ binary mask tubes, where $n$ represents the number of qualifying categories in the video.

\subsection{Dataset Statistics}

The statistics of the proposed benchmark are shown in Table~\ref{tab:statistics}.
In our proposed benchmark, we collect a total of 369 videos, with an average of 270 frames per video (99,549 frames in the whole set).
%
%
The videos contain annotations for 30 semantic categories. 
All videos are used in the validation process.

We also report the source distribution of the source videos. 
As shown in Figure~\ref{fig:dataset_source}(a), we gather a total of 369 videos, with 63.7$\%$ originating from the LVOS v2~\cite{lvosv2} dataset, 27.9$\%$ from MOSE~\cite{mose}, and 8.4$\%$ from DAVIS~\cite{davis}. 
The LVOS v2 dataset predominantly features longer videos, whereas shorter videos are more prevalent in MOSE and DAVIS.
This distribution results in the average length of our collected videos falling at a moderate level in Table~\ref{tab:statistics}. Figure~\ref{fig:dataset_source}(b) and (c) show the word cloud illustrating the proportion of frames for each category in our dataset, as well as the number of clips and frames per category. 
The categories with a relatively higher number of clips in our dataset are ``person,'' ``dog,'' and ``cat.'' 
Meanwhile, some categories, despite having fewer clips, contain a large number of frames, such as ``kite,'' ``surfboard,'' and ``aeroplane.'' 
Figure~\ref{fig:dataset_source}(d) presents multiple test cases from the benchmark, including category examples such as ``giraffe,'' ``motorcycle,'' and ``dog.'' 
%
%
Given a support image (e.g., one containing a ``dog'') and its corresponding semantic mask, the model must segment the same semantic regions in the video (\emph{e.g.}, track all semantic pixels of ``dog'' across the video). 
This process assesses the model's ability to transfer semantic understanding from static images to dynamic videos under one-shot learning conditions.

\begin{figure*}[t]
   \begin{center}
      \includegraphics[width=1\linewidth]{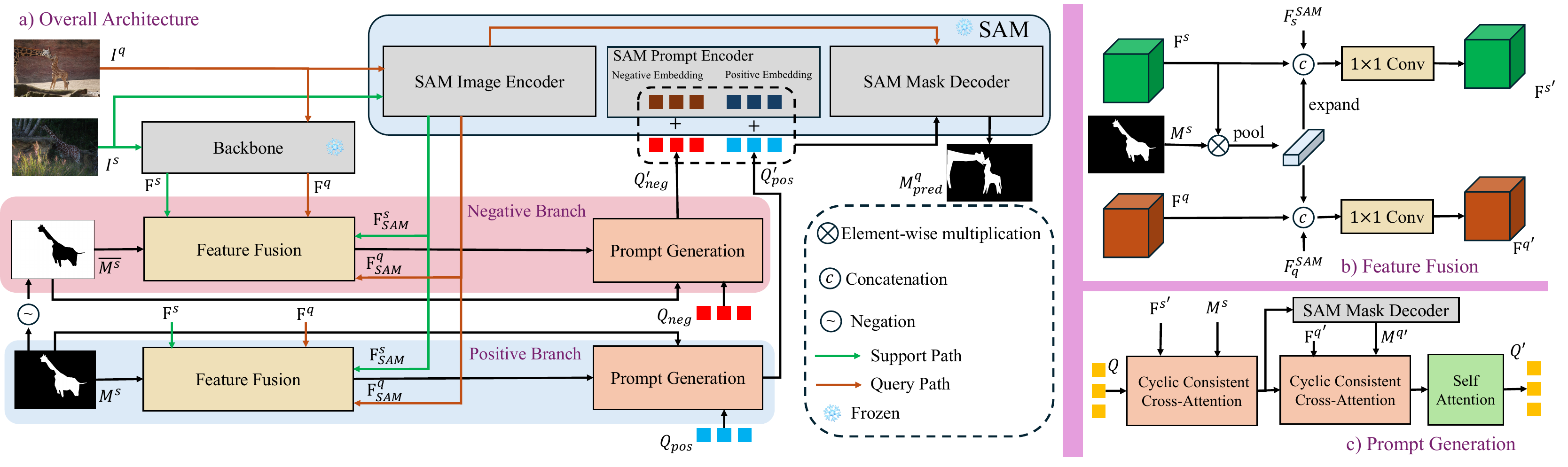}
   \end{center}
   \vspace{-5mm}
      \caption{Overview of the proposed DC-SAM framework. We use positive and negative branches to generate respective prompts, thereby refining the scope of the final generated mask. Additionally, we incorporate SAM features during the prompt generation process to better capture the characteristics of SAM, resulting in more accurate prompt boundaries. During the prompt generation process, we introduce cyclic consistent cross-attention to filter out non-cycle-consistent feature points, enhancing the precision of the prompts.}
      \vspace{-3mm}
   \label{fig:overview}
\end{figure*}

\subsection{IC-VOS Benchmark}

We evaluate state-of-the-art few-shot segmentation methods on the proposed dataset to establish the IC-VOS benchmark.
The evaluated methods encompass large-model-based approaches (\emph{e.g.,} PerSAM~\cite{persam}, Matcher~\cite{matcher}, VRP-SAM~\cite{vrpsam}) and traditional few-shot segmentation models (PFENet~\cite{pfenet}, HDMNet~\cite{hdmnet}, AMNet~\cite{amnet}).
Since these methods are primarily designed for image-level tasks, we implement modifications to ensure fair and meaningful comparisons.
Specifically, these models are integrated with SAM2~\cite{ravi2024sam2} to leverage its mask propagation capabilities.
The specific modifications are detailed below.

For large-model-based approaches (\emph{e.g.,} PerSAM, Matcher, VRP-SAM), the intermediate outputs of the first frame (\emph{e.g.,} prompts) are fed into SAM2 for inference and propagation.
For conventional few-shot segmentation methods (\emph{e.g.,} PFENet, HDMNet, AMNet), the final binary mask prediction from the first frame is used by SAM2 for propagation.

Trainable models (\emph{i.e.,} VRP-SAM, PFENet, HDMNet, AMNet) are retrained on the COCO few-shot segmentation dataset (all four folds combined) for 50 epochs, following the learning rates specified in their original works.
For non-trainable models (\emph{i.e.,} Matcher, PerSAM), inference is performed directly without additional training or fine-tuning.

\begin{algorithm}[h]
\small
\caption{DC-SAM}
\label{alg:dc-sam}
\KwIn{
    \begin{itemize}
        \item Support image $I^s$ with mask $M^s$
        \item Query image $I^q$
        \item Feature extractor $f_\theta$
        \item SAM Image Encoder $f_\theta^{SAM}$
        \item SAM Positive/Negative Embeddings $E_{pos}$, $E_{neg}$
    \end{itemize}
}
\KwOut{Predicted mask $M_{pred}^q$}
\vspace{0.2cm}
\textbf{Step 1: Feature Extraction}
\Indp
    \begin{align}
        &F^s \gets f_\theta(I^s),\quad\text{Support Feature:} \\
        &F^q \gets f_\theta(I^q),\quad\text{Query Feature:} \\
        &F_{SAM}^s \gets f_\theta^{SAM}(I^s),\quad\text{Support SAM Feature} \\
        &F_{SAM}^q \gets f_\theta^{SAM}(I^q)\quad\text{Query SAM Feature}
    \end{align}
\Indm \\

\textbf{Step 2: Feature Fusion}
\Indp
    \begin{align}
        F^{s'} &\gets \text{Conv}_{1\times1}\left(\text{Concat}(F^s, M^s \otimes F^s, F_{SAM}^s)\right) \\
        F^{q'} &\gets \text{Conv}_{1\times1}\left(\text{Concat}(F^q, M^s \otimes F^s, F_{SAM}^q)\right)
    \end{align}
\Indm

\textbf{Step 3: Prompt Initialization}
\Indp
    \begin{align}
        Q_{pos} &\gets \text{RandomInit}(), \\
        Q_{neg} &\gets \text{RandomInit}()
    \end{align}
\Indm

\textbf{Step 4: Mediate Prompt Generation}
\Indp
    \begin{align}
        Q_{pos}^{med} &\gets \text{QCycAttn}(Q_{pos}, F^{s'}, M^s), \\
        Q_{neg}^{med} &\gets \text{QCycAttn}(Q_{neg}, F^{s'}, M^s)
    \end{align}
\Indm

\textbf{Step 5: Pseudo Query Mask Generation}
\Indp
    \begin{equation}
        M^{q'}\gets \text{SAMMaskDecoder}(Q_{pos}^{med}+E_{pos},Q_{neg}^{med}+E_{neg})
    \end{equation}
\Indm

\textbf{Step 6: Final Prompt Refinement}
\Indp
    \begin{align}
        Q_{pos}^{'} &\gets \text{SelfAttn}(\text{QCycAttn}(Q_{pos}^{med}, F^{q'}, M^{q'})), \\
        Q_{neg}^{'} &\gets \text{SelfAttn}(\text{QCycAttn}(Q_{neg}^{med}, F^{q'}, M^{q'}))
    \end{align}
\Indm

\textbf{Step 7: Final Mask Prediction}
\Indp
    \begin{equation}
        M_{pred}^q\gets \text{SAMMaskDecoder}(Q_{pos}^{'}+E_{pos},Q_{neg}^{'}+E_{neg})
    \end{equation}
\Indm

\Return $M_{pred}^q$\;
\end{algorithm}

\section{Proposed Method}



\subsection{Overview}

As illustrated in Figure~\ref{fig:overview} and Figure~\ref{fig:sam2_overview}, our proposed method builds upon SAM and SAM2.
SAM is an interactive segmentation model capable of generating high-quality masks from given prompts.
The model consists of three primary components: the image encoder, the prompt encoder, and the mask decoder.
The image encoder extracts high-quality features for segmentation, and the prompt encoder processes visual prompts (such as boxes, points, masks, or text) to generate the corresponding tokens. 
The mask decoder receives the image features and the tokens encoded from the prompts, utilizing a bidirectional transformer decoder to generate semantic masks.
SAM2 further improves SAM by extending interactive segmentation to video.
It also has a memory module, including a memory encoder and a memory bank, to track each pixel in time.
Notably, both SAM and SAM2 share an identical prompt encoder design.
Our approach efficiently leverages the inherent capabilities of each SAM component during the prompt generation process, particularly for positive and negative points.
Thus, our method can be applied to both architectures, resulting in unified modeling of in-context segmentation in both images and videos.

As depicted in Figure~\ref{fig:overview} (a), our proposed DC-SAM primarily models the SAM prompt encoder by integrating in-context information into the prompt generation process.
%
%
Specifically, it employs dual branches to generate positive and negative prompts.
For each branch, the prompt generation process is bifurcated into feature fusion and consistent prompt generation. 
This integration ensures that the prompt generation process accounts for SAM features while leveraging complementary insights from mask priors. 
To refine query-based prompt generation, we introduce a new cycle-consistent cross-attention mechanism to exclude regions irrelevant to the desired prompts. 
This mechanism is applied twice to produce refined prompts.
%

\subsection{Feature Extraction and Fusion}
\label{subsec:feature_extraction}


Similar to the prior design in few-shot segmentation~\cite{pfenet}, we extract features from both the query image and the support image before the prompt generation process.
Specifically, given the input support image $I^s$ and the query image $I^q$, the backbone initially extracts features from these images.
The backbone network can be the pre-trained ResNet-50~\cite{he2016deep}, VGG-16~\cite{simonyan2014very}, or DINOv2~\cite{oquab2023dinov2}.
Following the conventional design~\cite{pfenet,cyctr}, we utilize intermediate layer features (from the third and fourth stages of the backbone) and apply convolution operations to reduce dimensions, yielding initial features $F_b^s$ and $F_b^q$.
The prior masks are computed using the support image's corresponding mask $M^s$ and high-level features (from the fifth stage).
This mask determines the pixel-wise similarity between the query and support features, retaining the maximum similarity at each pixel and normalizing the similarity map to the range $[0, 1]$ using min-max normalization.
%


As illustrated in Figure~\ref{fig:overview}, we concatenate the mask-averaged support feature with the query and support features, as well as the features extracted by SAM ($F_S^s$ and $F_S^q$), which are the same size as the query and support features.
These concatenated features are then processed by a $1 \times 1$ convolution before being fed into the transformer, ultimately yielding $F^s$ and $F^q$:
\begin{align}
    F^{s'} &= \text{Conv}_{1\times1}\left(\text{Concat}(F^s, M^s \otimes F^s, F_{SAM}^s)\right), \\
    F^{q'} &= \text{Conv}_{1\times1}\left(\text{Concat}(F^q, M^s \otimes F^s, F_{SAM}^q)\right).
\end{align}
By integrating SAM's features, both $F^s$ and $F^q$ become better aligned for the prompt generation process.


%
%
%
%

\subsection{Consistent Prompt Generation}
\label{subsec:prompt_generation}
%
Motivated by CycTR~\cite{cyctr}, our method includes the cyclic consistent cross-attention mechanism and the self-attention mechanism.
%
%
However, unlike CycTR, which primarily enforces pixel-level attention consistency between query and support features, our goal is to ensure cyclic consistency for visual prompts compatible with SAM.
%
%

Specifically, the initialized random visual prompts are regarded as query $Q$ in Figure~\ref{fig:overview}(c).
We compute cross-attention between query and support features, with the fused features $F^{s'}$ and $F^{q'}$ serving as the key and value inputs, respectively, as shown in Figure~\ref{fig:overview}(c).
First, we compute the affinity map $A=\frac{QK^T}{\sqrt{d}}, A\in\mathbb{R}^{N\times H_s W_s}$ to evaluate the similarity between the query and each support feature. 
For each pixel $j$ in the support features, where $j\in \{0, 1, \cdots, H_s W_s-1\}$, $H_s$ and $W_s$ represent the height and width of the support features respectively,  the affinity map $A$ is used to select the index $i^{*}$ of the query with the highest similarity:
\begin{equation}
    i^* = \mathop{\arg\max}\limits_{i} A(i, j),
\end{equation}
where $i\in \{0, 1, \cdots, N-1\}$ and $N$ is the number of specified queries. Applying the same method, we identify the pixel index $j^{*}$ of the support feature that has the highest similarity to the selected query:
\begin{equation}
    j^* = \mathop{\arg\max}\limits_{j} A(i^*, j).
\end{equation}

\begin{figure}[tp]
   \begin{center}
    \includegraphics[width=1\linewidth]{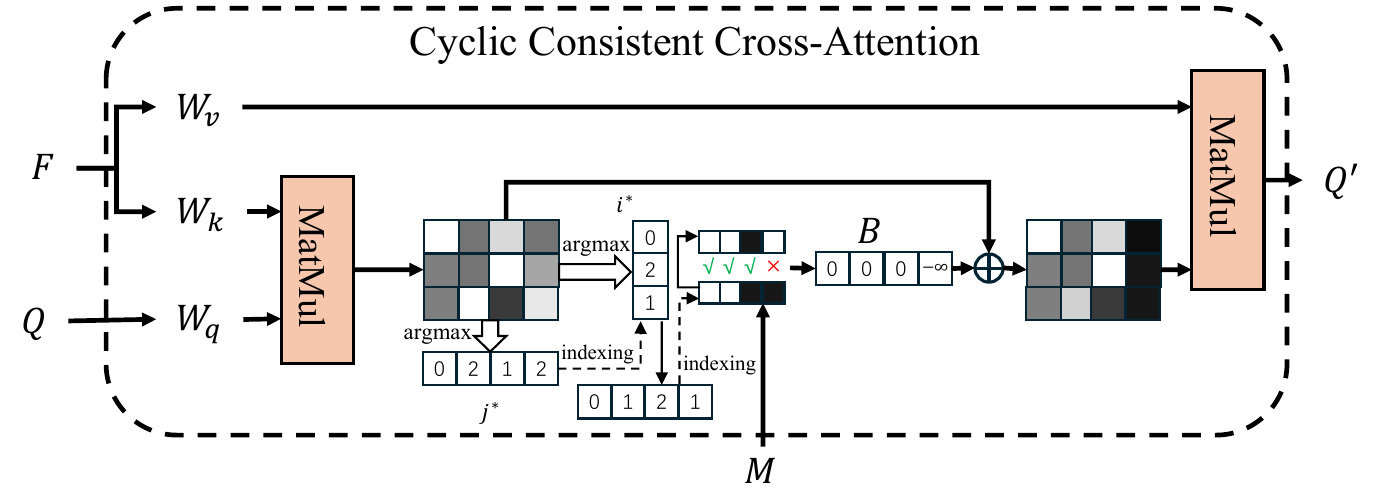}
   \end{center}
   \vspace{-4mm}
      \caption{Illustration of our proposed cyclic consistent cross-attention mechanism. This figure shows the version applied to query features with one head. The ``Cyc'' operation represents the process described in Equation~\ref{eq:cyc}, which ultimately generates a bias to filter out features that are not cycle-consistent.}
      \label{fig:cyc_consistency}
   \label{fig3}
   \vspace{-2mm}
\end{figure}

\begin{figure}[tp]
   \begin{center}
      \includegraphics[width=1\linewidth]{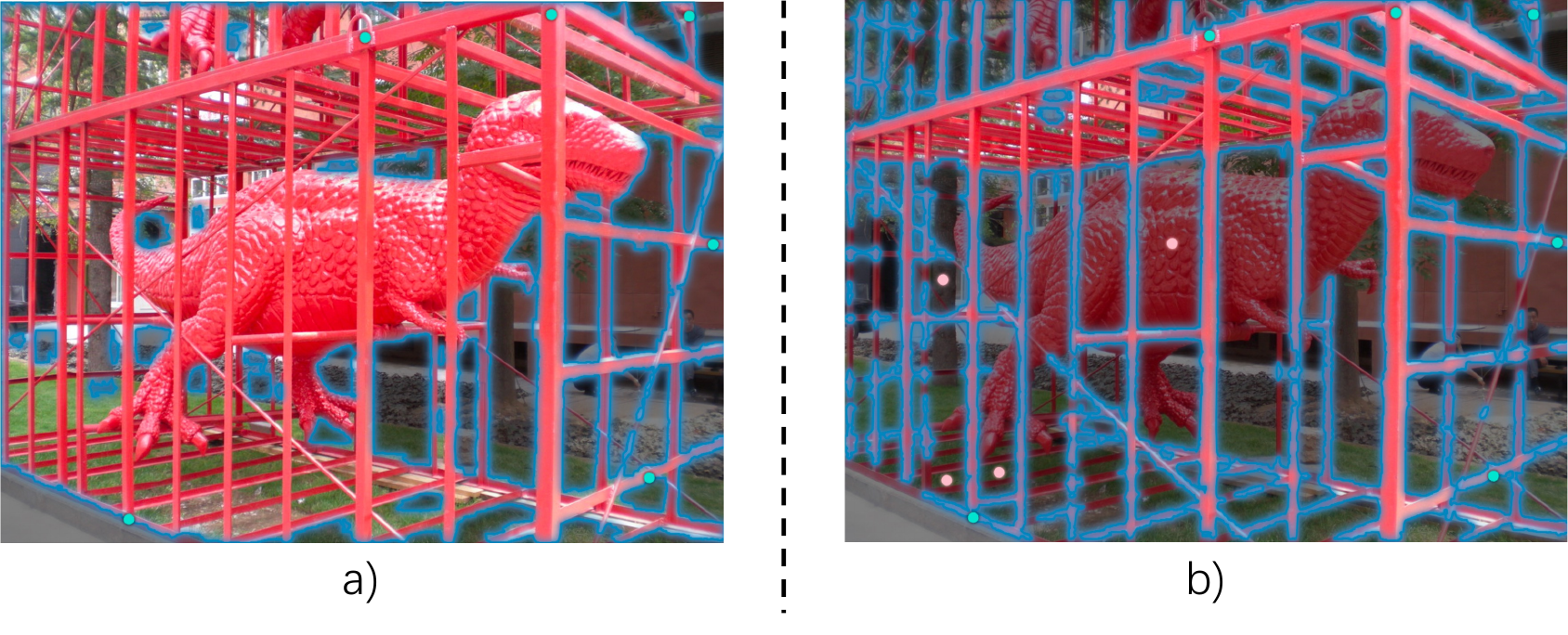}
   \end{center}
   \vspace{-5mm}
      \caption{Comparison of SAM segmentation results with and without negative prompts. (a) Segmentation of the cage using only positive prompts. (b) Segmentation of the cage using both positive and negative prompts. Although not achieving optimal segmentation results, adding negative prompts allowed for better differentiation between the background, the dinosaur, and the cage, resulting in a significantly improved result. }
   \label{fig:posneg}
   \vspace{-3mm}
\end{figure}

\begin{figure*}[tp]
   \begin{center}
      \includegraphics[width=1\linewidth]{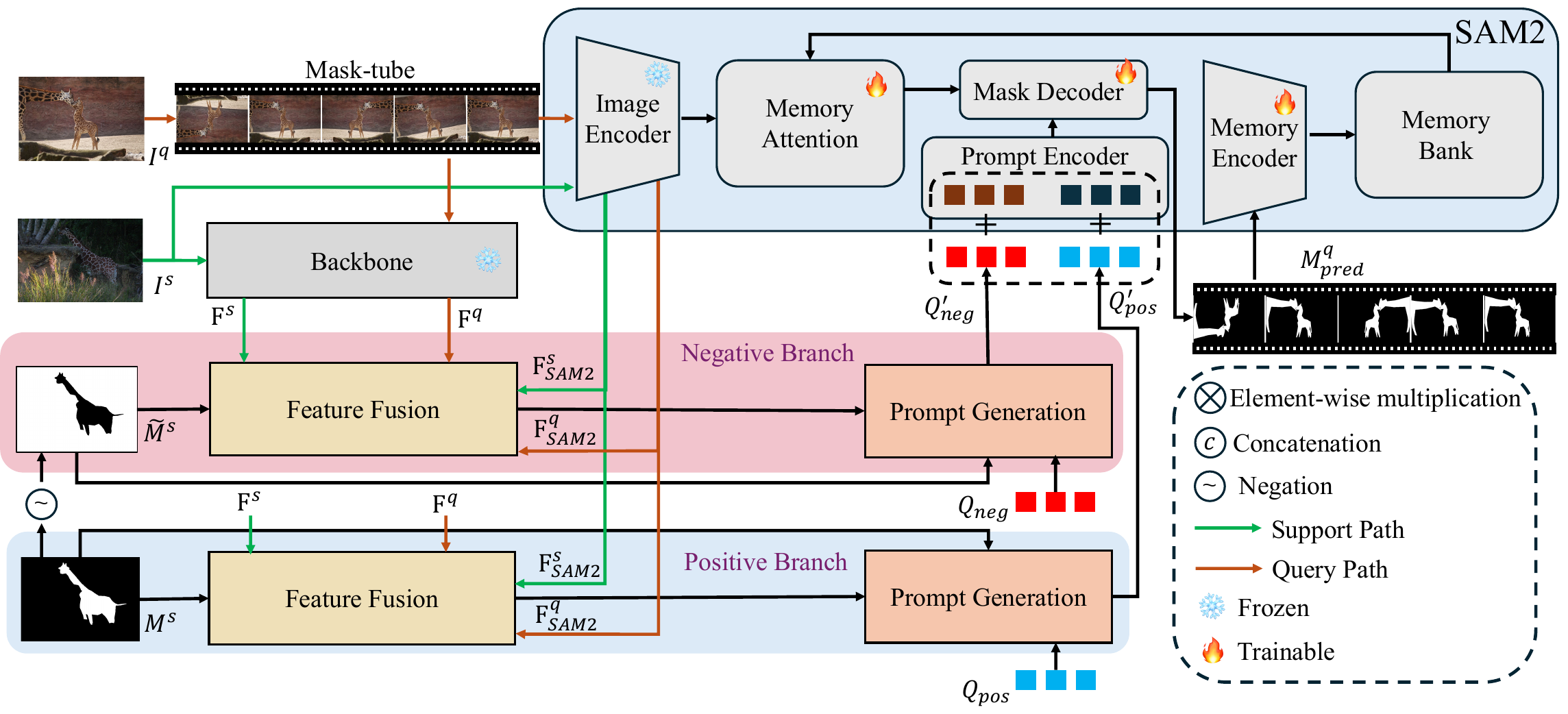}
   \end{center}
   \vspace{-2mm}
      \caption{Illustration of our proposed DC-SAM framework with SAM2. Unlike the image-level framework, we train the entire model for the video to acquire the image-to-video prompt ability. We apply different data augmentation techniques to the query image, and the augmented images compose a mask tube for training.}
   \label{fig:sam2_overview}
   \vspace{-3mm}
\end{figure*}

Given the flattened mask $M_s\in\mathbb{R}^{H_s W_s}$ corresponding to the support image, cyclic consistency can be determined using the identified pixel indices. 
Cyclic consistency is satisfied when $M_{s(j)}=M_{s(j^{*})}$, and we incorporate this constraint into the cross-attention calculation to encourage the model to learn more cyclically consistent results. 
Specifically, we calculate a bias $B\in\mathbb{R}^{H_s W_s}$ using the following formula:
\begin{equation}
    B_j = \begin{cases}
    0, & \text{if } M_{s(j)} = M_{s(j^*)} \\
    -\infty, & \text{if } M_{s(j)} \neq M_{s(j^*)}
    \end{cases},
    \label{eq:cyc}
\end{equation}
In this manner, the attention weights for cyclically inconsistent features are set to zero, thereby effectively filtering out features that should not be included in the prompt. 
For each query $Q_i\in\mathbb{R}^d$, we compute the result of the cross-attention as follows: 
\begin{equation}
    QCycAttn(Q_i, K, V) = \text{Softmax}(A_{i} + B) \cdot V,
\end{equation}
where $A$ denotes the affinity matrix without masking.
Since we do not have the masks for the query features as in Equation~\ref{eq:cyc}, it is challenging to apply the aforementioned Cyclic Consistent Cross-Attention. 
Alternatively, the queries can be fed into the SAM mask decoder to generate a pseudo-mask $\hat{M}^q$ for the query features.
This also allows us to apply the Cyclic Consistent Cross-Attention to the query features, as shown in the middle of Figure~\ref{fig:overview}(c). 
The above process is repeated with the refine query $Q'$ and query feature $F^{q'}$ as input.
After these two Cyclic Consistent Cross-Attention layers, we further perform a self-attention operation on all queries, forcing the global view of query features.
This finally yields the generated prompts $P$. 


%

\subsection{Dual Prompt-aware Mask Prediction}
\label{subsec:mask_prediction}

%
SAM uses positive and negative prompts during the training process to achieve flexible fine-grained mask control.
As shown in Figure~\ref{fig:posneg}, using only positive sample points as prompts results in segmentation results with coarse and imprecise mask edges. 
However, the addition of a negative sample point as a prompt significantly improves the mask edges.
Thus, we leverage this inherent characteristic of SAM by combining positive and negative prompts to achieve superior segmentation results.

Specifically, we employ two branches to generate positive and negative prompts, respectively. For the positive branch, we use the support mask $M^s$ to participate during the feature extraction process for the target category and then to generate the positive prompt $P_{pos}$. 
For the negative branch, we invert the support mask to obtain the background mask $\overline{M^s}=1 - M^s$, which indicates the region where the negative prompt should be located, and use it to generate the negative prompt $P_{neg}$.
Furthermore, we utilize SAM's inherent method of encoding positive and negative sampling points, by leveraging SAM's prompt encoder to enable it to perceive the differences between positive and negative prompts. 
In particular, SAM uses positive and negative embeddings, $E_{pos}$ and $E_{neg}$, to annotate the input prompts. 
We adopt a similar approach by labeling the generated positive and negative samples, which result in $P_{pos}^{'} = P_{pos} + E_{pos}$ and $P_{neg}^{'} = P_{neg} + E_{neg}$. These labeled prompts can be fed into SAM's mask decoder to obtain the predicted mask $M_{pred}^{q}$.

More importantly, our work can be easily extended to SAM2 for in-context video segmentation, since we only improve the prompt encoder parts.
As shown in Figure~\ref{fig:sam2_overview}, we design a mask tube prediction for the input query video during training.
The mask tube is created by stacking enhanced image masks into a video, encapsulating semantic-level spatiotemporal information.
Each mask tube represents the semantics traced in the temporal domain.
Without bells and whistles, this simple design further boosts the performance of in-context video segmentation on our proposed dual consistency baseline.
In particular, we jointly fine-tune the SAM2 decoder, memory modules, and our proposed dual consistency prompt generation module.

\subsection{Optimization and Inference}
\label{sec:training_inference}

\noindent
\textbf{Training.} 
We adopt the common segmentation training setup for both SAM and SAM2.
We directly use two loss functions to guide the segmentation training process. 
The Binary Cross-Entropy Loss (BCE Loss) supervises the pixel-level binary mask output by the model, as follows:
\begin{equation}
    \mathcal{L}_{BCE} = -\frac{1}{N}\sum_{i=1}^N [y_i \log(p_i) + (1-y_i) \log(1-p_i)],
\end{equation}
where $N$ denotes the total number of pixels in the image, $y_i$ represents the true label value of pixel $i$, and $p_i$ is the probability value predicted by the SAM model at pixel $i$. 
Additionally, the Dice Loss is employed to provide additional context for pixel-level segmentation, by addressing class imbalance issues, as follows:
\begin{equation}
    \mathcal{L}_{Dice} = 1-\frac{2\cdot\sum_{i=1}^N(p_i\cdot y_i)}{\sum_{i=1}^N p_i^2 + \sum_{i=1}^N y_i^2}
\end{equation}

Consequently, the total loss function used in our model is a combination of the BCE Loss and the Dice Loss:
\begin{equation}
    \mathcal{L} = \mathcal{L}_{BCE} + \mathcal{L}_{Dice}
\end{equation}

\noindent
\textbf{Inference.}
We process the query and support images and the support mask using DC-SAM for the \emph{image in-context segmentation} task. 
The resulting positive and negative prompts are then combined with the pre-trained positive and negative embeddings to generate point prompts. 
These combined embeddings, enriched with contextual information, are then fed into the SAM/SAM2 mask decoder to produce final masks, ensuring accurate and contextually relevant segmentation outcomes.

However, for the \emph{video in-context segmentation} task, we apply DC-SAM to the first frame of the input video using the same procedure as in the image setting, thereby obtaining the mask output for the first frame. 
This mask is then inputted into SAM2 to generate memory embeddings, which are propagated across subsequent video frames utilizing the SAM2 mask decoder.
Thus, we can obtain the final mask tube of the entire video.

\section{Experiments}
\label{sec:exp}

    


    




    
 

\noindent\textbf{Datasets.} 
We first evaluate the proposed method on the IC-VOS benchmark. 
In addition, we use the setting of~\cite{vrpsam} and evaluate the proposed DC-SAM on two other widely adopted datasets: PASCAL-$5^i$~\cite{shaban2017one} and COCO-$20^i$~\cite{nguyen2019feature}. 
The PASCAL-$5^i$ is derived from PASCAL VOC 2012~\cite{everingham2010pascal} with additional annotations from SDS~\cite{hariharan2011semantic}, encompassing 20 classes. COCO-$20^i$ is based on MS-COCO~\cite{lin2014microsoft} and includes 80 categories. 
For each dataset, we perform cross-validation by evenly dividing all classes into four folds. 
We adhere to the same class splits specified in~\cite{nguyen2019feature,shaban2017one} for PASCAL-$5^i$ and COCO-$20^i$, respectively. 
Specifically, three folds are used for training, while the remaining fold is reserved for testing. 


\noindent\textbf{Evaluation Metrics.} 
We use the mean intersection over union (mIoU) as our primary evaluation metric.
%
We denote $\text{mIoU}=\frac{1}{C} \sum_{i=1}^{C} \text{IoU}_i$, where $C$ represents the total number of classes to be evaluated in each fold, and $\text{IoU}_i$ represents the Intersection-over-Union for the $i$-th class.
%
%
This metric disregards class-specific distinctions and computes the average value across all classes.
For the IC-VOS benchmark, two commonly-used metrics were adopted: region similarity  ($\mathcal{J}$) and contour accuracy ($\mathcal{F}$). 
These metrics are also used by the video sources of our dataset: MOSE~\cite{mose}, LVOS v2~\cite{lvosv2}, and DAVIS~\cite{davis}. 
We calculate the mean of these values as the final score.


\noindent\textbf{Image Benchmarks.} We implement  DC-SAM using the PyTorch~\cite{paszke2019pytorch} framework, employing various encoders to generate prior masks for support and query images, including VGG-16~\cite{simonyan2014very}, ResNet-50~\cite{he2016deep}, Swin-B~\cite{liu2021swin}, and DINO v2-B/14~\cite{oquab2023dinov2}. 
We used AdamW~\cite{loshchilovdecoupled} as an optimizer and employed a cosine learning rate decay strategy for training. 
For the COCO-20$^i$ dataset, the model is trained for 50 epochs with a learning rate of $1 \times 10^{-4}$ and a batch size of 8.
For the PASCAL-$5^i$ dataset, training was conducted for 100 epochs with a learning rate of $2 \times 10^{-4}$ and a batch size of 8. 
A weight decay of $1 \times 10^{-5}$ was applied to both datasets during training.
The input image size was fixed at $512 \times 512$, and no data augmentation techniques were applied.
Note that due to the patch size of 14 in DINO v2-B/14, we scale the image size to $896 \times 896$ when using DINO v2-B/14 as the prior masks generator to ensure that its output size matches the feature size of SAM. 
The number of queries in both the positive and negative branches was set to 25.

\noindent\textbf{IC-VOS Benchmark.} For the IC-VOS benchmark, our prompt generator is connected with SAM2 to provide prompts for the first image, enabling SAM2 to propagate these prompts throughout the video sequence.
The structure and optimizer of the prompt generator are identical to those used in the image benchmark. 
The differences primarily involve the number of training iterations, training methods, and learning rate configurations.
In the first training step, we pre-train our entire framework on COCO images, specifically training the prompt generator and the SAM2 decoder while freezing all other parameters. 
This step involves $40,000$ iterations with a learning rate of $1 \times 10^{-4}$. 
In the second step, we generate mask tubes from the query image using various image augmentation techniques to enhance the performance of our framework in the video domain. 
We also unfreeze the memory encoder and memory attention parameters, and this step involves $10,000$  iterations with a learning rate of $1 \times 10^{-5}$.

\subsection{Main Results}
\label{sec:main_results}

\noindent\textbf{Comparison with SAM-based models and visual foundation models.} 
We evaluate our proposed DC-SAM against other SAM-based methods, including PerSAM~\cite{persam}, Matcher~\cite{matcher}, and VRP-SAM~\cite{vrpsam}, on the COCO-$20^i$ dataset.
As shown in Table~\ref{tab:foundation},
DC-SAM outperforms other current SAM-based methods.
We also compare our approach with methods based on visual foundational models, such as Painter~\cite{painter} and SegGPT~\cite{seggpt}. 
%
%
Table~\ref{tab:foundation} shows that the method based on DC-SAM and DINO v2-B surpasses SegGPT by 6\%, wdespite SegGPT being trained on large-scale in-domain datasets.
These results demonstrate the effectiveness of our approach with less data tuning.


\noindent\textbf{Comparison with few-shot segmentation methods.} We present the evaluation results of our model and recent few-shot segmentation methods on the COCO-$20^i$ and PASCAL-$5^i$ datasets in Table~\ref{tab:total}.
We use different backbones to generate prior masks, including VGG16~\cite{simonyan2014very}, ResNet50~\cite{he2016deep}, and DINO v2-B/14~\cite{oquab2023dinov2}, each representing different architectures and feature extraction capabilities. 
As illustrated in Table~\ref{tab:total}, DC-SAM performs favorably across various backbone configurations.
Specifically, our model outperforms existing state-of-the-art few-shot segmentation models in every fold of each dataset and under all backbone settings.
These results highlight significant performance advantages and the exceptional generalization capabilities of DC-SAM in different scenarios and categories. 
%


\begin{table}[t]
\caption{Comparison with other few-shot segmentation models with foundational models on the COCO-20$20^i$ dataset. Methods marked with * indicate using external data. Methods marked with a $\dag$ symbol indicate SAM-based models.}
\vspace{-1mm}
  \centering
  \small
  \begin{tabular}{l|cccc|c} 
    \toprule
    Method & F-0 & F-1 & F-2 & F-3 & Means\\ \midrule
    Painter*~\cite{painter} & 31.2 & 35.3 & 33.5 & 32.4 & 33.1 \\
    SegGPT*~\cite{seggpt} & 56.3 & 57.4 & 58.9 & 51.7 & 56.1 \\
    PerSAM-F~\cite{persam}$\dag$ & 22.3 & 24.0 & 23.4 & 24.1 & 23.5 \\
    Matcher~\cite{matcher}$\dag$ & 52.7 & 53.5 & 52,6 & 52.1 & 52.7 \\ 
    \midrule
    VRP-SAM~\cite{vrpsam}$\dag$ & & & & & \\
    \quad - ResNet50 & 48.1 & 55.8 & 60.0 & 51.6 & 53.9 \\
    \quad - DINO v2-B & \textbf{56.8} & \underline{61.0} & \underline{64.2} & \underline{59.7} & \underline{60.4} \\ \midrule
    DC-SAM$\dag$ & & & & & \\
    \quad - ResNet50 & 50.4 & 56.0 & 61.0 & 54.4 & 55.5 \\
    \quad - DINO v2-B & \textbf{56.8} & \textbf{62.0} & \textbf{67.3} & \textbf{61.9} & \textbf{62.0} \\
    \bottomrule
  \end{tabular}
  \label{tab:foundation}
  \vspace{-2mm}
\end{table}


\begin{table*}[t]\footnotesize
 \caption{Comparison with the state-of-the-art few-shot segmentation methods on COCO-$20^i$~\cite{nguyen2019feature} and PASCAL-$5^i$~\cite{everingham2010pascal}. The best results are highlighted in \textbf{bold}, while the second-best results are \underline{underlined}.}
 \vspace{-1mm}
 \centering
 \renewcommand\arraystretch{1.2}
 \setlength{\tabcolsep}{3.4mm}{
 \resizebox{0.9\linewidth}{!}{
\begin{tabular}{l|c|ccccc|ccccc}
\hline
\multirow{2}{*}{\textbf{Method}} &
  \multirow{2}{*}{\begin{tabular}[c]{@{}c@{}}\textbf{Image} \\ \textbf{encoder}\end{tabular}} &
  \multicolumn{5}{c|}{\textbf{COCO-20$^i$}} &
  \multicolumn{5}{c}{\textbf{PASCAL-5$^i$}} \\ \cline{3-12} 
 & & F-0 & F-1 & F-2 & \multicolumn{1}{c|}{F-3} & Mean & F-0 & F-1 & F-2 & \multicolumn{1}{c|}{F-3} &  Mean \\ \hline
PFENet~\cite{pfenet} & \multirow{4}{*}{VGG-16} & 35.4 & 38.1 & 36.8 & \multicolumn{1}{c|}{34.7} & 36.3 & 56.9 & 68.2 & 54.5 & \multicolumn{1}{c|}{52.4} & 58.0 \\
BAM~\cite{bam} &  & 36.4 & 47.1 & 43.3 & \multicolumn{1}{c|}{41.7} & 42.1 & 63.2 & 70.8 & 66.1 & \multicolumn{1}{c|}{57.5} & 64.4 \\
HDMNet~\cite{hdmnet} &  & 40.7 &50.6 &48.2 &\multicolumn{1}{c|}{44.0} &45.9 &64.8 &71.4 &67.7 &\multicolumn{1}{c|}{56.4} &65.1 \\
 VRP-SAM~\cite{vrpsam} & &43.6 &51.7 &50.0 &\multicolumn{1}{c|}{46.5} &48.0 &70.0 &74.7 &68.3 &\multicolumn{1}{c|}{61.9} &68.7 \\
 \hline
 DC-SAM & VGG-16 & 44.7 & 50.2 & 59.1 & \multicolumn{1}{c|}{50.6} & 51.2 & 71.7 & 77.2 & 69.0 & \multicolumn{1}{c|}{63.8} & 70.4 \\\midrule
PFENet~\cite{pfenet} &\multirow{12}{*}{ResNet-50} &36.5 &38.6 &34.5 &\multicolumn{1}{c|}{33.8} &35.8 & 61.7 &69.5 & 55.4 &\multicolumn{1}{c|}{56.3} & 60.8 \\
HSNet~\cite{hsnet} &  &36.3 & 43.1 & 38.7 &\multicolumn{1}{c|}{38.7} &39.2 &64.3 &70.7 &60.3 &\multicolumn{1}{c|}{60.5} & 64.0 \\
CyCTR~\cite{cyctr} &  & 38.9 &43.0 &39.6 &\multicolumn{1}{c|}{39.8} &40.3 &65.7 &71.0 &59.5 &\multicolumn{1}{c|}{59.7} & 64.0 \\
SSP~\cite{ssp} & & 35.5 &39.6 &37.9 &\multicolumn{1}{c|}{36.7} &37.4 &60.5 &67.8 &66.4 &\multicolumn{1}{c|}{51.0} & 61.4 \\
NTRENet~\cite{ntre} & & 36.8 &42.6 &39.9 &\multicolumn{1}{c|}{37.9} &39.3 &65.4 &72.3 &59.4 &\multicolumn{1}{c|}{59.8} & 64.2 \\
DPCN~\cite{dpcn}& & 42.0 &47.0 &43.3 &\multicolumn{1}{c|}{39.7} &43.0 &65.7 &71.6 &69.1 &\multicolumn{1}{c|}{60.6} & 66.7 \\
VAT~\cite{vat} & & 39.0 &43.8 &42.6 &\multicolumn{1}{c|}{39.7} &41.3 &67.6 &72.0 &62.3 &\multicolumn{1}{c|}{60.1} & 65.5 \\

BAM~\cite{bam} & & 39.4 & 49.9 & 46.2 &\multicolumn{1}{c|}{45.2} &45.2 &69.0 &73.6 & 67.6 & \multicolumn{1}{c|}{61.1} & 67.8 \\
HDMNet~\cite{hdmnet} & & 43.8 & 55.3 & 51.6 & \multicolumn{1}{c|}{49.4} &50.0 &71.0 & 75.4 & 68.9 & \multicolumn{1}{c|}{62.1} & 69.4 \\
AMNet~\cite{amnet} & & 44.9 & 55.8 & 52.7 & \multicolumn{1}{c|}{50.6} & 51.0 & 71.1 & 75.9 & 69.7 & \multicolumn{1}{c|}{63.7} & 70.1  \\
ABCB~\cite{abcb} & & 44.2 & 54.0 & 52.1 & \multicolumn{1}{c|}{49.8} & 50.0 & 72.9 & 76.0 & 69.5 & \multicolumn{1}{c|}{64.0} & 70.6 \\
VRP-SAM~\cite{vrpsam} & &48.1 &55.8 &60.0 &\multicolumn{1}{c|}{51.6} &53.9 &73.9 &78.3 &\underline{70.6} &\multicolumn{1}{c|}{65.0} &\underline{71.9} \\
\hline
DC-SAM &  \multirow{2}{*}{ResNet-50} & \textbf{50.4} & \underline{56.0} & \underline{61.0} & \multicolumn{1}{c|}{\underline{54.4}} & \underline{55.5} & \underline{74.8} & \textbf{79.1} & \textbf{71.4} & \multicolumn{1}{c|}{\underline{66.5}} & \underline{73.0} \\ 
DC-SAM (SAM2) & & \underline{49.7} & \textbf{56.4} & \textbf{63.1} & \multicolumn{1}{c|}{\textbf{56.2}} & \textbf{56.4} & \textbf{77.4} & \underline{78.5} & 70.5 & \multicolumn{1}{c|}{\textbf{69.4}} & \textbf{74.0}\\
\bottomrule
\end{tabular}}}
\label{tab:total}
\vspace{-2mm}
\end{table*}

\begin{table}[tbp]
\caption{Results on IC-VOS benchmark. Bold and underlined texts indicate the best and second-best results, respectively.}
\vspace{-1mm}
  \centering
  \small
  \begin{tabular}{l|ccc} 
    \toprule
    Method & $\mathcal{J}$ & $\mathcal{F}$ & $\mathcal{J\&F}$\\ \midrule
    PerSAM~\cite{persam} + SAM2 & 32.23 & 36.81 & 34.52 \\ 
    PerSAM-F~\cite{persam} + SAM2 & 31.52 & 36.83 & 34.18 \\ 
    Matcher~\cite{matcher} + SAM2 & 26.88 & 24.00 & 20.44 \\ 
    VRP-SAM~\cite{vrpsam} + SAM2 & 50.77 & 56.30 & 53.53 \\\midrule
    PFENet~\cite{pfenet} + SAM2 & \underline{62.07} & \underline{66.45} & \underline{64.26} \\
    HDMNet~\cite{hdmnet} + SAM2 & 53.07 & 57.49 & 55.28 \\
    AMNet~\cite{amnet} + SAM2 & 53.51 & 58.36 & 55.94 \\ \midrule
    DC-SAM & \textbf{68.38} & \textbf{74.65} & \textbf{71.52} \\
    \bottomrule
  \end{tabular}
  \label{tab:v_quan}
  \vspace{-3mm}
\end{table}

\noindent
\textbf{Quantitative Results on the IC-VOS Benchmark.} As shown in Table~\ref{tab:v_quan}, DC-SAM surpasses all other evaluated models, achieving a $\mathcal{J}\&\mathcal{F}$ score of $71.52$ on the IC-VOS benchmark. 
This represents an 11.3\% improvement over the second-best performing method, PFENet~\cite{pfenet}.
%
Our model can effectively segment the corresponding semantic regions in videos based on a few categorical examples.

\begin{figure*}[tp]
   \begin{center}
      \includegraphics[width=0.8\linewidth]{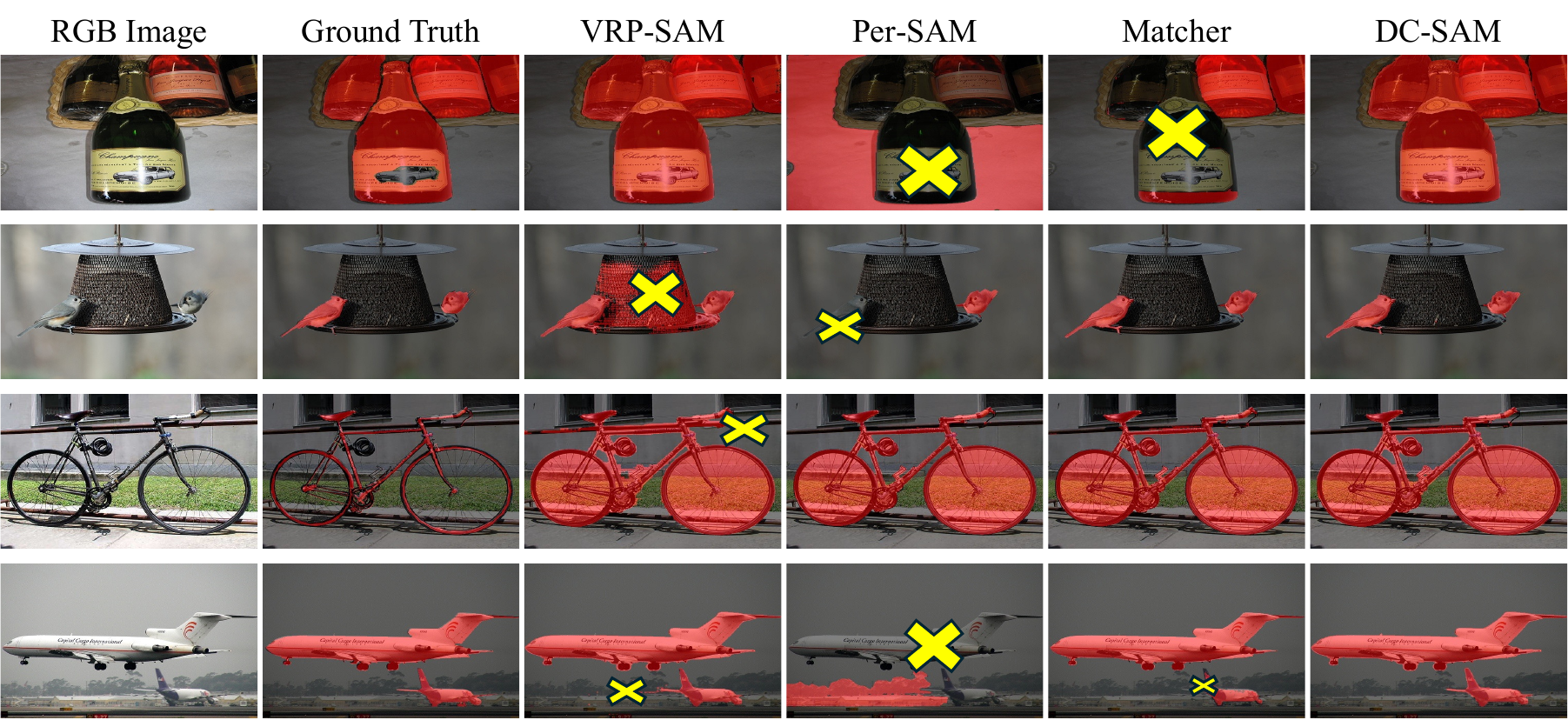}
   \end{center}
   \vspace{-3mm}
      \caption{Comparison of segmentation results from different methods on the PASCAL-$5^i$. Each row displays an RGB image along with its corresponding ground truth segmentation and the results of the four methods. Notable errors are marked with a yellow ``$\times$''.}
   \label{fig:vis}
   \vspace{-3mm}
\end{figure*}

\begin{figure*}[t]
   \begin{center}
      \includegraphics[width=0.8\linewidth]{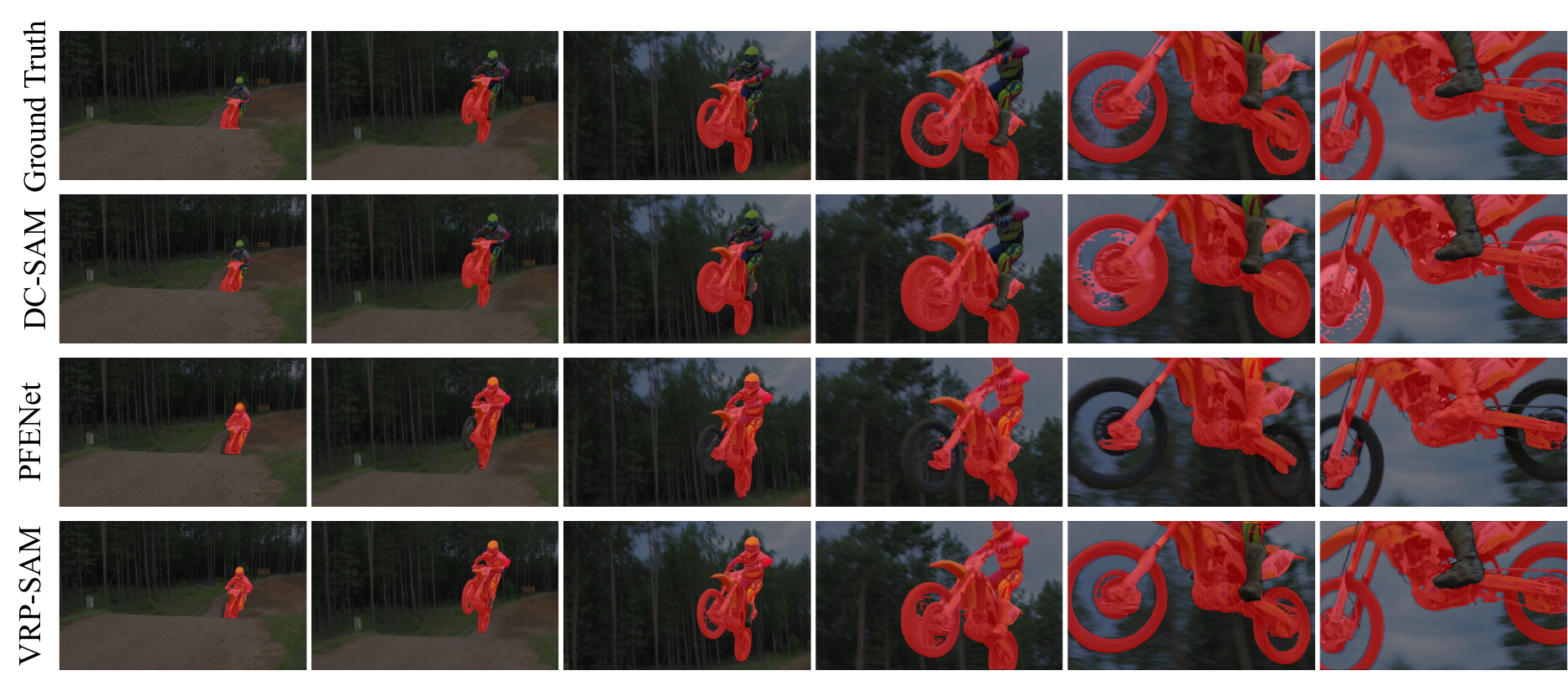}
   \end{center}
   \vspace{-2mm}
      \caption{Visual comparisons of our proposed model with PFENet and VRP-SAM on our proposed benchmark. The support mask in the video indicates the semantic category of motorcycle, and all three models share the same support image and mask.}
   \label{fig:video_vis}
   \vspace{-2mm}
\end{figure*}

\subsection{Visual Comparisons}
\label{sec:visual_comparison}

\noindent\textbf{Images.}~Figure~\ref{fig:vis} shows sample segmentation results of the SAM-based methods on the PASCAL-$5^i$ dataset. 
%
%
For the ``bottle'' example shown in the first row, DC-SAM segments the objects accurately with the complete contours.  
For the ``bird'' example in the second row, DC-SAM segments the objects with fine details. 
Similarly, in the ``bicycle'' and ``aeroplane'' examples presented in the third and fourth rows, DC-SAM consistently performs effectively with no false positives in background regions and accurately captures the complex contours of target objects. 
Overall, DC-SAM can segment complex objects in the PASCAL-$5^i$ dataset with fine details. 
%
%

\noindent\textbf{Videos.}~Figure~\ref{fig:video_vis} presents sample results of DC-SAM against PFENet and VRP-SAM on few-shot video semantic segmentation using the proposed benchmark. 
We show the results from a single video clip in the dataset, and all three models receive the same support image and mask, indicating the target semantic category of motorcycle. 
DC-SAM accurately identifies the motorcycle and maintains good performance in subsequent segmentation. 
%
In contrast, PFENet overlooks the wheels of the motorcycle and also segments the person together with the motorcycle. Similarly, VRP-SAM segments both the person and the motorcycle in the earlier frames.
%
%
%

\begin{table}[tbp]
  \caption{Ablation study on each innovation of the model. We start with VRP-SAM~\cite{vrpsam} as the baseline and incrementally add our innovations on the PASCAL-$5^i$.}
  \vspace{-1mm}
  \centering
  \small
  \resizebox{1.0\linewidth}{!}{
  \begin{tabular}{l|cccc|cc} 
    \toprule
    Ablation & F-0 & F-1 & F-2 & F-3 & Means & $\Delta$\\ \midrule
    VRP-SAM~\cite{vrpsam} & 73.9 &78.3 &70.6 &65.0 &71.9 & 0 \\
    + Pos-Neg Branch & 74.0 & 78.5 & 70.3 & 65.5 & 72.1 & +0.2 \\
    + SAM Feature Fusion & \textbf{74.8} & \textbf{79.6} & \underline{70.7} & \underline{66.2} & \underline{72.8} & +0.6 \\
    + Cyclic Consistent & \textbf{74.8} & \underline{79.1} & \textbf{71.4} & \textbf{66.5} & \textbf{73.0} & +0.2 \\ 
    \bottomrule
  \end{tabular}
  }
  \label{tab:ablation}
  \vspace{-2mm}
\end{table}

\begin{figure}[t]
   \begin{center}
      \includegraphics[width=0.9\linewidth]{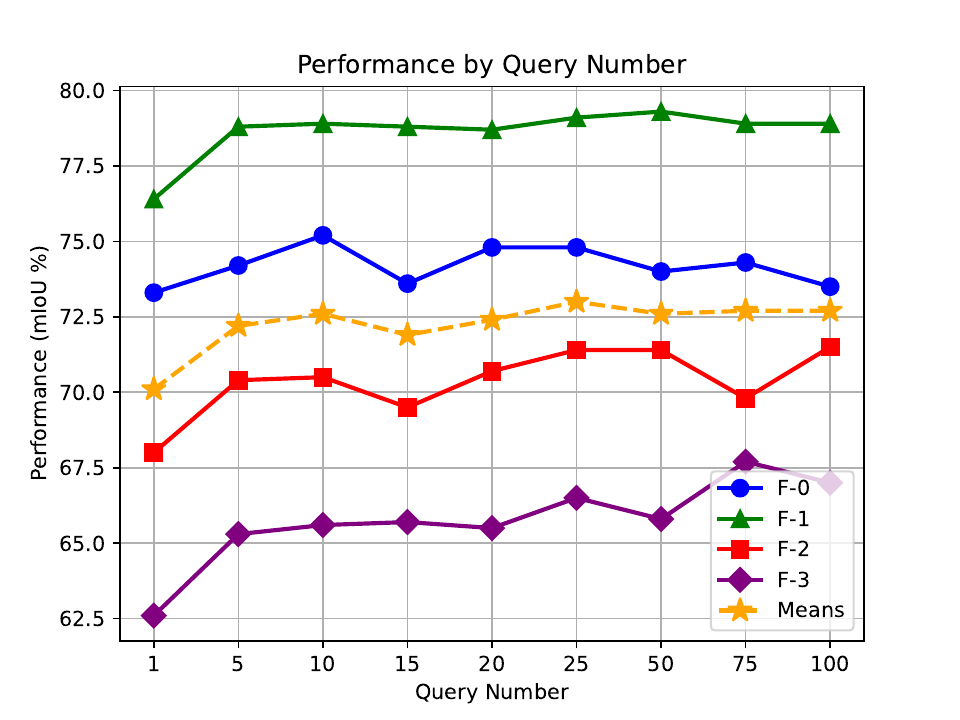}
   \end{center}
   \vspace{-5mm}
       \caption{Ablation study on the number of queries. The $x$-axis represents the number of queries in one branch. Note that since our DC-SAM consists of both positive and negative branches, the total number of queries is twice the number shown on the $x$-axis. The $y$-axis represents the model's performance. These experiments are conducted on the PASCAL-$5^i$ dataset.}
   \label{fig:query_num}
   \vspace{-2mm}
\end{figure}

\begin{table}[t]
\caption{Performance comparison of different pre-trained modules. During fine-tuning, the SAM2 decoder parameters were unfrozen and adjusted. The term ``w/o pre-trained'' indicates direct training using mask tubes over 50,000 iterations.}
\vspace{-1mm}
  \centering
  \small
  \begin{tabular}{c|ccc} 
    \toprule
    Ablation & $\mathcal{J}$ & $\mathcal{F}$ & $\mathcal{J\&F}$\\ \midrule
    w/o pre-trained & 58.72 & 64.37 & 61.54 \\ 
    Prompt Generator & 64.16 & 70.30 & 67.23 \\ 
    Prompt Generator and decoder & \textbf{67.57} & \textbf{73.62} & \textbf{70.59} \\ 
    \bottomrule
  \end{tabular}
  \label{tab:p_module_ablation}
\end{table}

\begin{table}[t]
\caption{Comparison of fine-tuning with different modules.}
\vspace{-2mm}
  \centering
  \small
  \begin{tabular}{cc|ccc} 
    \toprule
    decoder & Memory & $\mathcal{J}$ & $\mathcal{F}$ & $\mathcal{J\&F}$\\ \midrule
    $\checkmark$& & 67.57 & 73.62 & 70.59 \\ 
    & $\checkmark$ & 68.27 & 74.20 & 71.23 \\ 
    $\checkmark$ & $\checkmark$ & \textbf{68.38} & \textbf{74.65} & \textbf{71.52} \\ 
    \bottomrule
  \end{tabular}
  \label{tab:v_comp_ablation}
  \vspace{-2mm}
\end{table}

\begin{table}[t]
\caption{Comparison of fine-tuning with and without mask tubes. The total number of training iterations is 50k.}
\vspace{-2mm}
  \centering
  \small
  \begin{tabular}{c|ccc} 
    \toprule
    Ablation & $\mathcal{J}$ & $\mathcal{F}$ & $\mathcal{J\&F}$\\ \midrule
    w/o mask tube & 63.85 & 70.03 & 66.94 \\ 
    w/ mask tube & \textbf{67.57} & \textbf{73.62} & \textbf{70.59} \\ 
    \bottomrule
  \end{tabular}
  \label{tab:masktube_ablation}
\end{table}

\subsection{Ablation Studies on DC-SAM}

\noindent\textbf{Ablation on Each Component.}
%
Table~\ref{tab:ablation} shows the effectiveness of each component of the proposed DC-SAM with ResNet50~\cite{he2016deep} as the backbone network in the PASCAL-$5^i$ dataset. Note that we begin by using VRP-SAM~\cite{vrpsam} as the baseline and progressively integrate DC-SAM.
Upon the introduction of positive and negative branches, the model has the ability to refine segmentation outcomes by leveraging positive and negative prompts, thus enhancing overall performance across each fold.
During feature extraction and fusion, the incorporation of the SAM feature makes the prompt generation progress more aligned. 
%
%
Finally, as shown in the last row, adding prompt consistency for each branch further leads to improvement over various folders.
%
%
This guides the model to focus more on the critical areas that require prompts, thus achieving precise segmentation.

\noindent\textbf{Query Number Ablation.} We also explore the effect of varying the number of queries $Q$ on the DC-SAM. 
As shown in Figure~\ref{fig:query_num}, the model's performance gradually improves as the number of queries in a single branch increases from 1. 
However, as the number of queries increases, the model's performance exhibits fluctuations, with improvements in some folds and declines in others. 
Overall, the model achieves the best average performance when the number of queries in a single branch is set to 25.

\subsection{Ablation Study on IC-VOS Benchmark}

\noindent\textbf{Pre-training.}~In Table~\ref{tab:p_module_ablation}, we show the performance of models trained using different approaches: (1) training the prompt generator and the SAM2 decoder for 50k iterations with data augmentation to generate mask tubes; (2) training only the prompt generator during pre-training; (3) jointly training the prompt generator and SAM2 decoder during pre-training. 
For fair comparisons, all methods utilizing pre-trained models fine-tune both the prompt generator and the SAM2 decoder. 
Experimental results demonstrate that pre-training on images followed by fine-tuning with mask tubes achieves the best video segmentation performance. 
This is because image datasets provide greater diversity, facilitating faster convergence, while fine-tuning on video datasets enhances correspondence learning for mask tubes.
Furthermore, training the SAM2 decoder during the image pre-training phase significantly enhances the final video segmentation performance.

\noindent\textbf{Fine-tuning with Ablated Modules.}
As shown in Table~\ref{tab:v_comp_ablation}, we show the performance variations when fine-tuning different modules of SAM2, including the prompt generator. 
The experimental results indicate that fine-tuning the memory module (including the memory encoder and memory attention) yields greater improvements compared to fine-tuning only the decoder. Fine-tuning both modules together results in optimal model performance.

\noindent\textbf{Mask-tube Fine-tuning.}~In our proposed DC-SAM, we utilize mask tubes generated by data augmentation to fine-tune the entire model.
These mask tubes offer a more diverse set of training examples, thereby enhancing the model's generalization capability across various scenarios. 
As shown in Table~\ref{tab:masktube_ablation}, we compare the performance of fine-tuning directly with images versus fine-tuning with mask tubes. 
Because fine-tuning the memory encoder and memory attention yields substantial improvements for video segmentation, we only train the prompt generator and mask decoder in this comparison to isolate the impact of the mask tubes. 
These results demonstrate that fine-tuning the model with mask tubes significantly improves video segmentation performance, highlighting the effectiveness of this approach in capturing temporal dynamics and enhancing segmentation consistency.

\begin{table}[tp]
\caption{Evaluation of DC-SAM in few-shot scenarios on the PASCAL-$5^i$ dataset.}
\vspace{-2mm}
\label{tab:5shot}
\resizebox{0.50\textwidth}{!}{
  \centering
  \begin{tabular}{c|c|cccc|c} 
    \toprule
    Shots & Backbone & F-0 & F-1 & F-2 & F-3 & Means\\ \midrule
    1 & \multirow{3}{*}{VGG-16~\cite{simonyan2014very}} & 71.7 & 77.2 & 69.0 & 63.8 &70.4 \\
    \multirow{2}{*}{5} & & 76.9 & 79.6 & 71.4 & 69.3 & 74.3 \\
     & & (\textbf{+5.2}) & (\textbf{+2.4}) & (\textbf{+2.4}) & (\textbf{+5.5}) & (\textbf{+3.9}) \\\midrule
    1 & \multirow{3}{*}{ResNet-50~\cite{he2016deep}} & 74.8 & 79.1 & 71.4 & 66.5 & 73.0 \\
    \multirow{2}{*}{5} & & 78.2 & 81.4 & 72.7 & 73.8 & 76.5 \\ 
     & & (\textbf{+3.4}) & (\textbf{+2.3}) & (\textbf{+1.3}) & (\textbf{+7.3}) & (\textbf{+3.5}) \\ 
    \bottomrule
  \end{tabular}
  }
\end{table}

\subsection{More Analysis}

\begin{table}[tp]
\caption{Performance and SAM2 trainable parameters of different fine-tuning versions.}
\vspace{-1mm}
  \centering
  \small
  \begin{tabular}{ccc|cccc} 
    \toprule
    Decoder & Memory & LoRA & \# param & $\mathcal{J}$ & $\mathcal{F}$ & $\mathcal{J\&F}$\\ \midrule
    $\checkmark$& & & 3.95M & 67.57 & 73.62 & 70.59 \\ 
    & $\checkmark$ & & 7.31M & 68.27 & 74.20 & 71.23 \\ 
    $\checkmark$ & $\checkmark$ & & 11.26M & \textbf{68.38} & \textbf{74.65} & \textbf{71.52} \\ 
    $\checkmark$ & $\checkmark$ & $\checkmark$ & 0.11M & 66.86 & 72.72 & 69.79\\ 
    \bottomrule
  \end{tabular}
  \label{tab:lora}
\end{table}

\begin{figure*}[t]
   \begin{center}
      \includegraphics[width=0.95\linewidth]{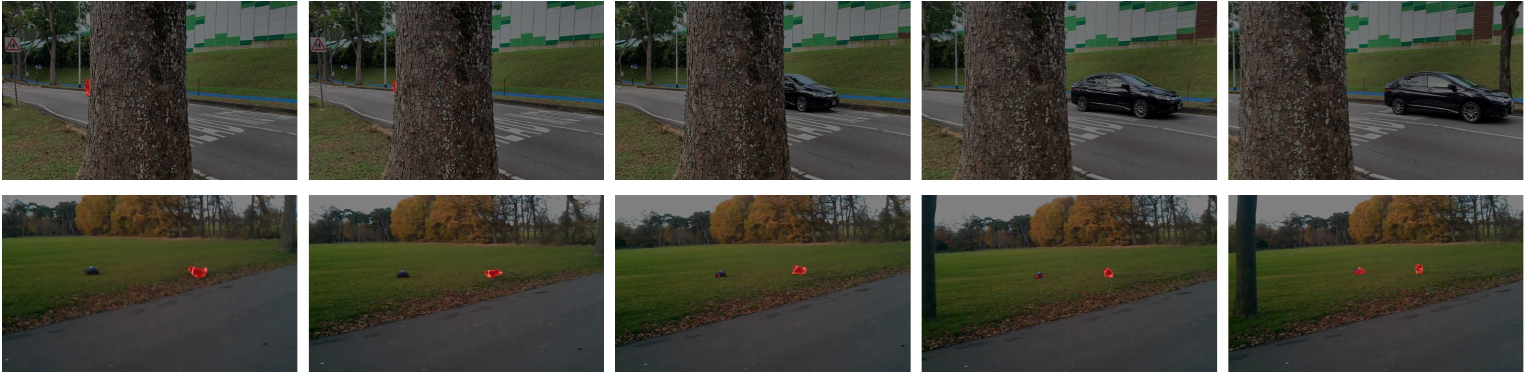}
   \end{center}
      \caption{Visualization of two failure cases of our proposed DC-SAM on IC-VOS. We still find missing matching objects due to the occlusion (in the top) and multiple instance inputs with the fast motion (in the bottom).}
   \label{fig:error_case}
\end{figure*}

\begin{table}[!htbp]
  \caption{Details of the data split for PASCAL-$5^i$ in the domain shift scenario. Each row represents non-overlapping classes in the training set, corresponding to the respective fold of COCO-$20^i$.}
  \vspace{-2mm}
  \centering
  \small
  \resizebox{0.4\textwidth}{!}{
  \begin{tabular}{c|c} 
    \toprule
    Fold & Test Classes\\ \midrule
    0 & Aeroplane, Boat, Chair, Dining table, Dog, Person \\
    1 & Horse, Sofa, Bicycle, Bus \\
    2 & Bird, Car, Potted plant, Sheep, Train, TV/monitor \\
    3 & Bottle, Cow, Cat, Motorbike \\
    \bottomrule
  \end{tabular}}
  \label{tab:non-overlap}
\end{table}

\begin{table}[!htbp]
  \caption{Generalization performance on the PASCAL-$5^i$ dataset using Mean IoU (\%).}
  \vspace{-2mm}
  \centering
  \small
  \begin{tabular}{lcc}
    \toprule
    Method & Image Encoder & Means\\ \midrule
    RPMM~\cite{rpmm} & \multirow{5}{*}{ResNet-50} & 49.6 \\
    PFENet~\cite{pfenet} & & 61.1 \\
    RePRI~\cite{repri} & & 63.2 \\
    VAT-HM~\cite{hm} & & 65.1 \\
    VRP-SAM~\cite{vrpsam} & &\underline{75.9}  \\ \midrule
    HSNet~\cite{hsnet} & \multirow{2}{*}{ResNet-101} & 64.1 \\
    DGPNet~\cite{dgpnet} & & 70.1 \\ \midrule
    FP-Trans~\cite{fptrans} & DeiT-B/16 & 69.7 \\ \midrule \midrule
    DC-SAM & ResNet-50 & \textbf{76.5}  \\
    \bottomrule
  \end{tabular}
  \label{tab:gene}
\end{table}

\begin{table}[h]
\caption{Results by the original SAM2 and the fine-tuned SAM2 on the proposed benchmark IC-VOS and the LVOS dataset. Both models were provided with the semantic/instance mask of the first frame of the video.}
  \centering
  \small
  \begin{tabular}{c|c|ccc} 
    \toprule
    Dataset & SAM2 Version & $\mathcal{J}$ & $\mathcal{F}$ & $\mathcal{J\&F}$\\ \midrule
    \multirow{2}{*}{IC-VOS} & original & 87.56 & 92.30 & 89.93 \\ 
    & fine-tuned & 89.07 & 94.41 & 91.74 \\ \midrule
    \multirow{2}{*}{LVOS v2~\cite{lvosv2}} & original & 75.18 & 81.97 & 78.58 \\ 
    & fine-tuned & 71.25 & 80.59 & 75.92 \\
    \bottomrule
  \end{tabular}
  \label{tab:performance_SAM2}
\end{table}

\begin{table}[h]
\caption{GFLOPs and learnable parameter analysis of our proposed model DC-SAM, VRP-SAM, and PFENet.}
\vspace{-2mm}
  \centering
  \small
  \begin{tabular}{c|cc} 
    \toprule
    Method & FLOPs (G) & Learnable Parameters\\ \midrule
    VRP-SAM~\cite{vrpsam} & 218.949 & 1.6M  \\ 
    PFENet~\cite{pfenet} & 207.582  & 10.8M \\ 
    DC-SAM & 278.97 & 1.9M \\
    \bottomrule
  \end{tabular}
  \label{tab:gflops}
\end{table}

\noindent\textbf{Extension to the Few-shot Setting.} To assess the performance of our proposed DC-SAM in the few-shot scenario, we evaluate the model under the 5-shot setting and compare the results with those from the 1-shot setting.
Table~\ref{tab:5shot} shows that the 5-shot model consistently surpasses the 1-shot model in performance across all folds.
The results indicate that our DC-SAM can be easily extended to the few-shot settings.

\noindent\textbf{Generalization Capability.}~To evaluate whether DC-SAM retains its generalization capability under domain-shift scenarios, we adopt the same experimental setups from prior studies~\cite{hsnet,pfenet,vrpsam}. 
Specifically, our model is trained on the COCO-$20^i$ dataset but tested on the PASCAL-$5^i$ dataset. 
As shown in Table~\ref{tab:non-overlap}, the categories for each fold of PASCAL-$5^i$ are adjusted based on the scheme in~\cite{vrpsam} to ensure that there is no overlap between the training and testing sets. 
As demonstrated in Table~\ref{tab:gene}, our model achieves state-of-the-art performance in generalization evaluation.
%

\noindent\textbf{LoRA Fine-Tuning of SAM2.}~We examine the effect of using Low-Rank Adaptation (LoRA)~\cite{lora} to fine-tune the SAM2 model, focusing on parameter reduction and final performance. 
As demonstrated in Table~\ref{tab:lora}, applying LoRA to fine-tune part of SAM2 reduces the parameter count by $99\%$ compared to full parameter fine-tuning, while retaining $97.6\%$ of the original model performance in terms of $\mathcal{J}\&\mathcal{F}$.
%

\noindent\textbf{Performance of SAM2 After Fine-Tuning.}~To evaluate whether fine-tuning SAM2 in our setup affects its original capabilities, we evaluate both the original and fine-tuned versions of SAM2 on our proposed benchmark and the LVOS v2~\cite{lvosv2} validation set. 
For each video clip, we use the mask of the first frame for the target semantic or instance segmentation, and we use SAM2 to infer the corresponding masks for all subsequent frames.
As demonstrated in Table~\ref{tab:performance_SAM2}, the fine-tuned SAM2 outperforms the original SAM2 on our proposed IC-VOS, demonstrating enhanced capabilities in semantic video segmentation. 
For the LVOS v2 video instance segmentation benchmark, the fine-tuned SAM2 shows a slight performance loss but retains $96.6\%$ of the original performance based on the $\mathcal{J}\&\mathcal{F}$ metric.
With more trained data or co-training with VOS data, DC-SAM is likely to achieve a better performance trade-off on both IC-VOS and VOS, which will be part of our future work. 
%

\noindent\textbf{Failure Cases.} Figure~\ref{fig:error_case} shows some failure segmentation results of DC-SAM. 
%
The first row shows a scenario where tracking fails due to occlusions,
leading to incorrect subsequent segmentation results from the propagation module of SAM2.  
The second row presents a case where the target semantic category is ``dog''. 
During fast motions involving the dog and a toy car, slight tracking errors occur despite initially accurate segmentation. 
In the third frame, a small portion of the mask erroneously tracks the wheel of the toy car, and the subsequent propagation process accumulates this error. 
%
%

\section{Conclusion}
\label{sec:conclusion}

In this paper, we present a prompt tuning-based method to adapt visual foundation models, SAM and SAM2, to better support in-context learners. 
The core idea is to leverage the features of the SAM prompt encoder to generate more fine-grained visual prompts with dual consistency.
Given the fused SAM features, we can use both positive and negative queries to generate visual prompts.
During the generation process, we design a cycle-consistent attention in each branch.
%
Furthermore, we propose a new dataset, IC-VOS, to benchmark existing representative methods combined with SAM2. 
Our proposed DC-SAM performs favorably against existing models on several few-shot segmentation benchmarks, even with SAM2. 
By adding a simple mask tube to DC-SAM, it also achieves state-of-the-art performance on the IC-VOS benchmark.
Extensive analysis shows the effectiveness, efficiency, and generalization of our approach.

\noindent
\textbf{Future Work.} 
The proposed model performs well in the few-shot image-to-image segmentation and the few-shot image-to-video segmentation task that we introduced. 
However, it still has some limitations, such as tracking errors due to occlusion, inaccurate prompts in the first frame, and large motion. 
We will address these issues in the future work.
\ifCLASSOPTIONcaptionsoff
  \newpage
\fi



{
\bibliographystyle{IEEEtran}
\bibliography{IEEEabrv,egbib}
}

\clearpage

\begin{figure*}[bp]
   \begin{center}
      \includegraphics[width=0.85\linewidth]{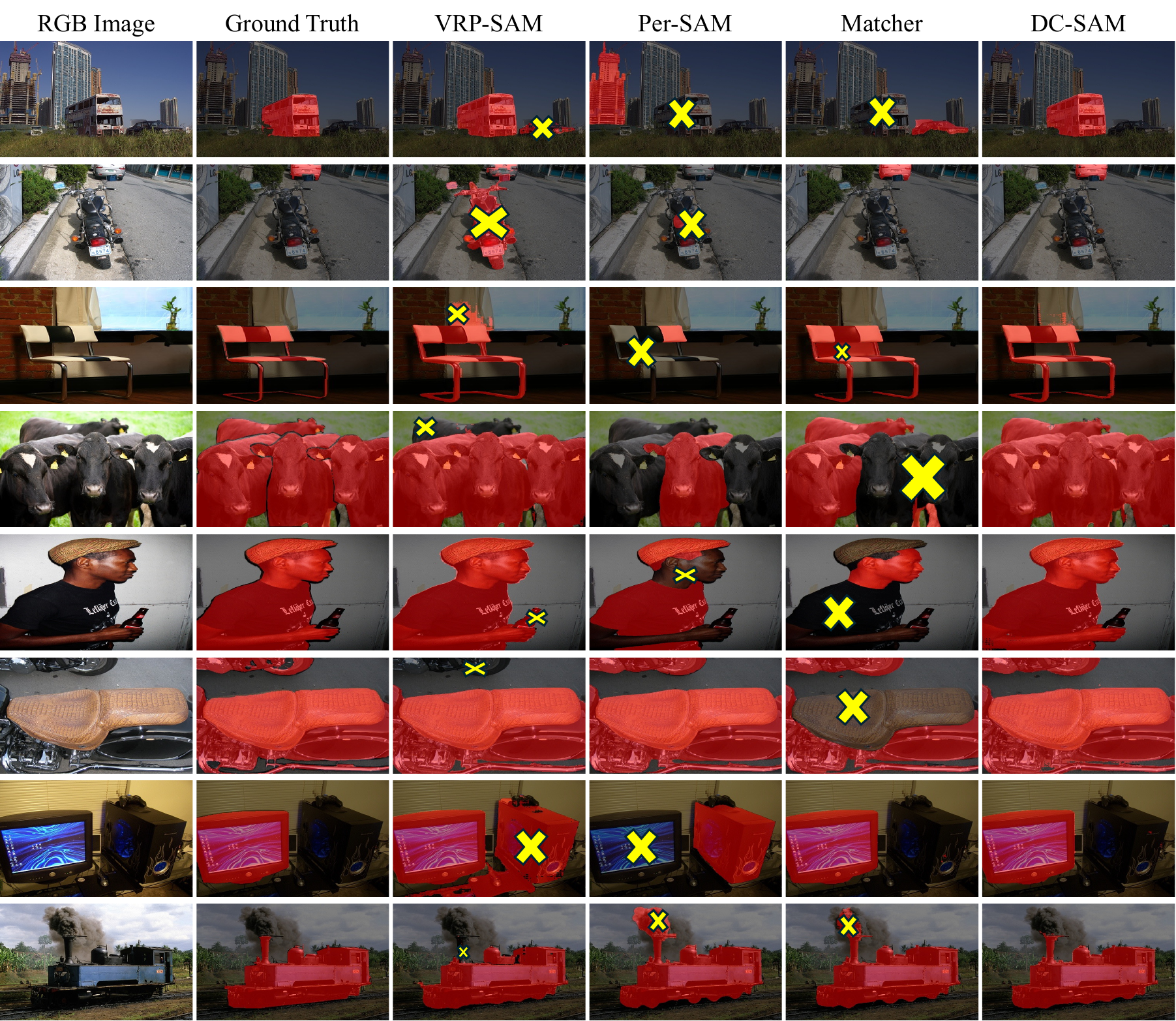}
   \end{center}
   \vspace{-3mm}
      \caption{Comparison of one-shot segmentation results on the PASCAL-$5^i$ dataset.}
   \label{fig:vis}
   \vspace{-3mm}
\end{figure*}

This supplementary material systematically demonstrates the extensive application effects of DC-SAM in image-to-image in-context segmentation and image-to-video in-context segmentation. To comprehensively validate the effectiveness and robustness of the method, we have carefully selected 20 representative image cases (covering various target categories) and 8 typical video sequences (including challenging scenarios such as dynamic target tracking and complex background changes) for visualization. By comparing the output results of DC-SAM with those of other state-of-the-art methods, the technical advantages of our approach in detail preservation and accurate prompting are intuitively highlighted. These rich visual examples not only corroborate the quantitative analysis conclusions presented in the main text but also provide multidimensional empirical references. They can be read in conjunction with the methodology section of the main text to gain a more comprehensive understanding.

\begin{figure*}[t]
   \begin{center}
      \includegraphics[width=0.9\linewidth]{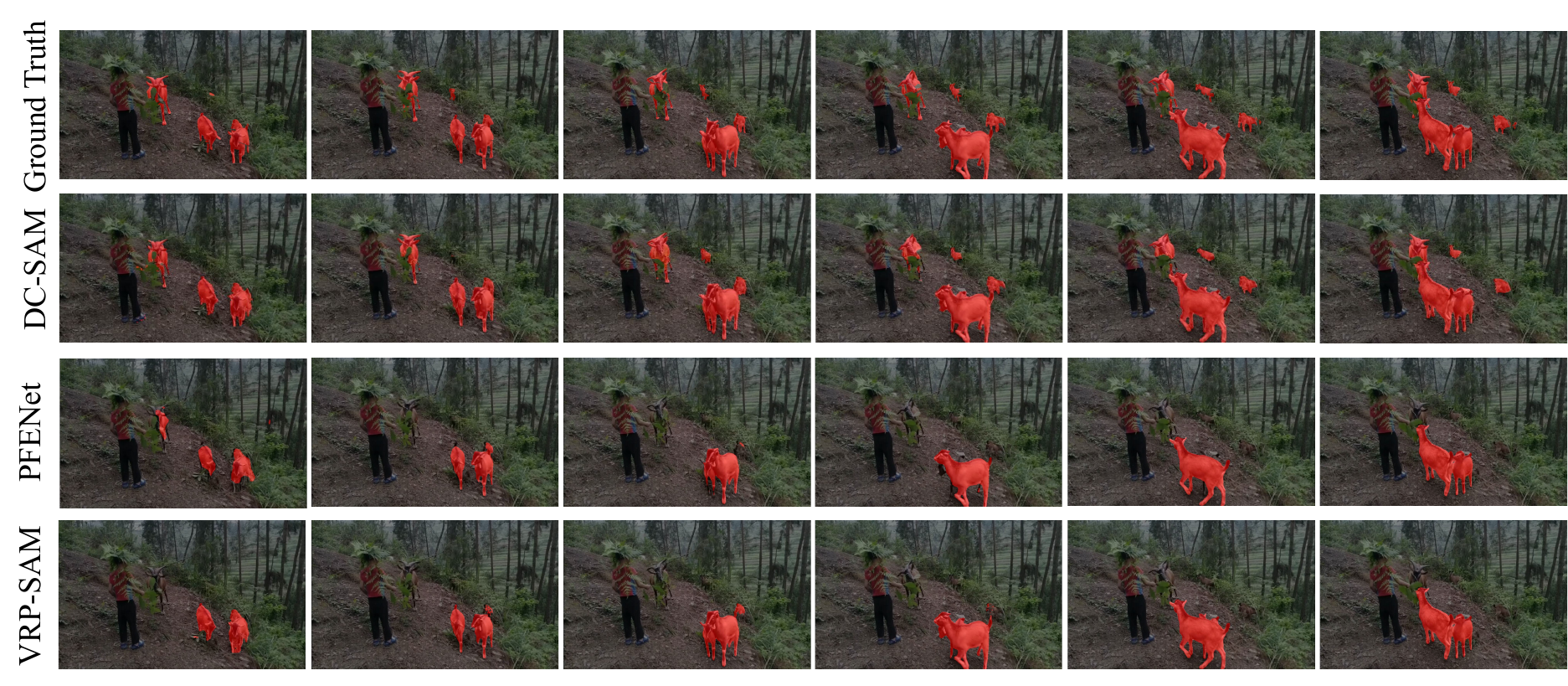}
   \end{center}
   \vspace{-3mm}
      \caption{Comparison of semantic segmentation results for the ``Sheep'' category on the IC-VOS dataset.}
   \label{fig:video_vis2}
   \vspace{-3mm}
\end{figure*}

\begin{figure*}[t]
   \begin{center}
      \includegraphics[width=0.9\linewidth]{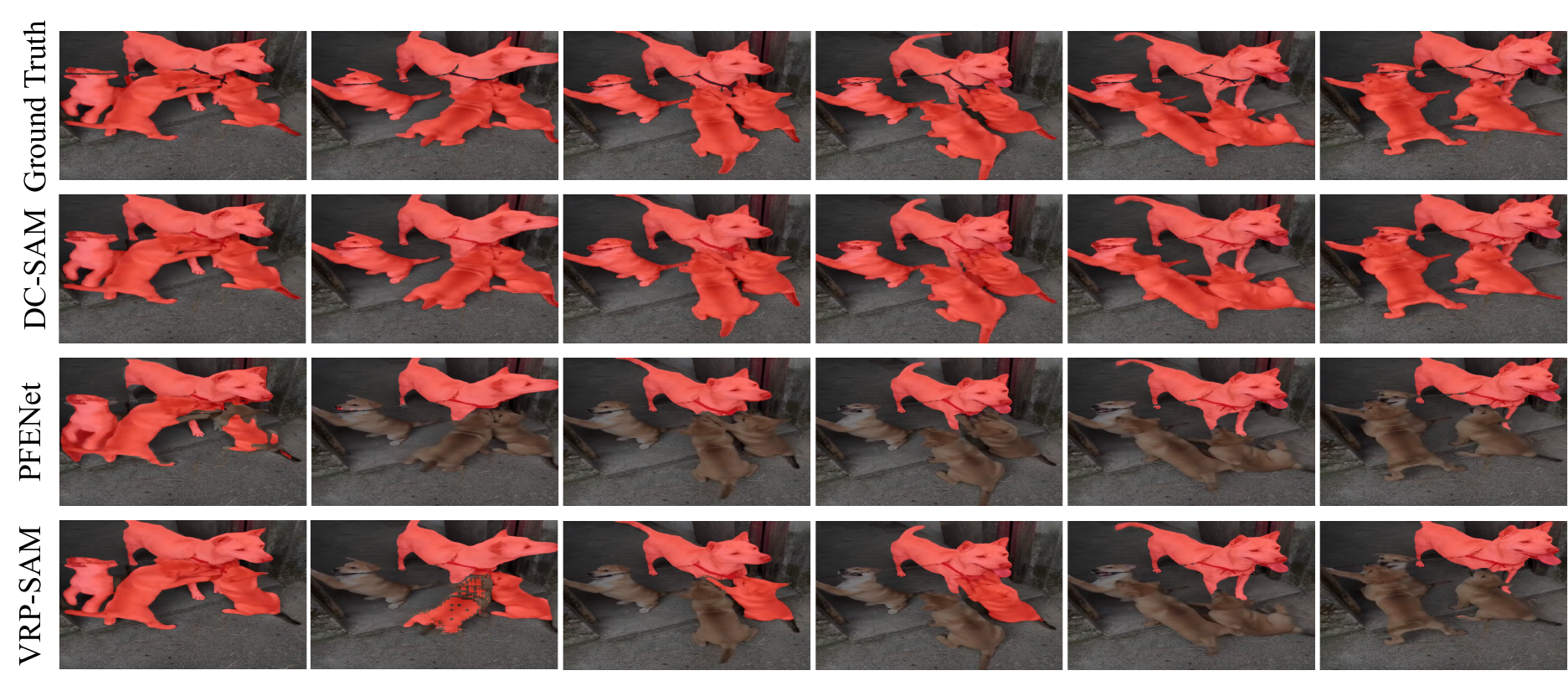}
   \end{center}
   \vspace{-3mm}
      \caption{Comparison of semantic segmentation results for the ``Dog'' category on the IC-VOS dataset.}
   \label{fig:video_vis3}
   \vspace{-3mm}
\end{figure*}

\begin{figure*}[t]
   \begin{center}
      \includegraphics[width=0.9\linewidth]{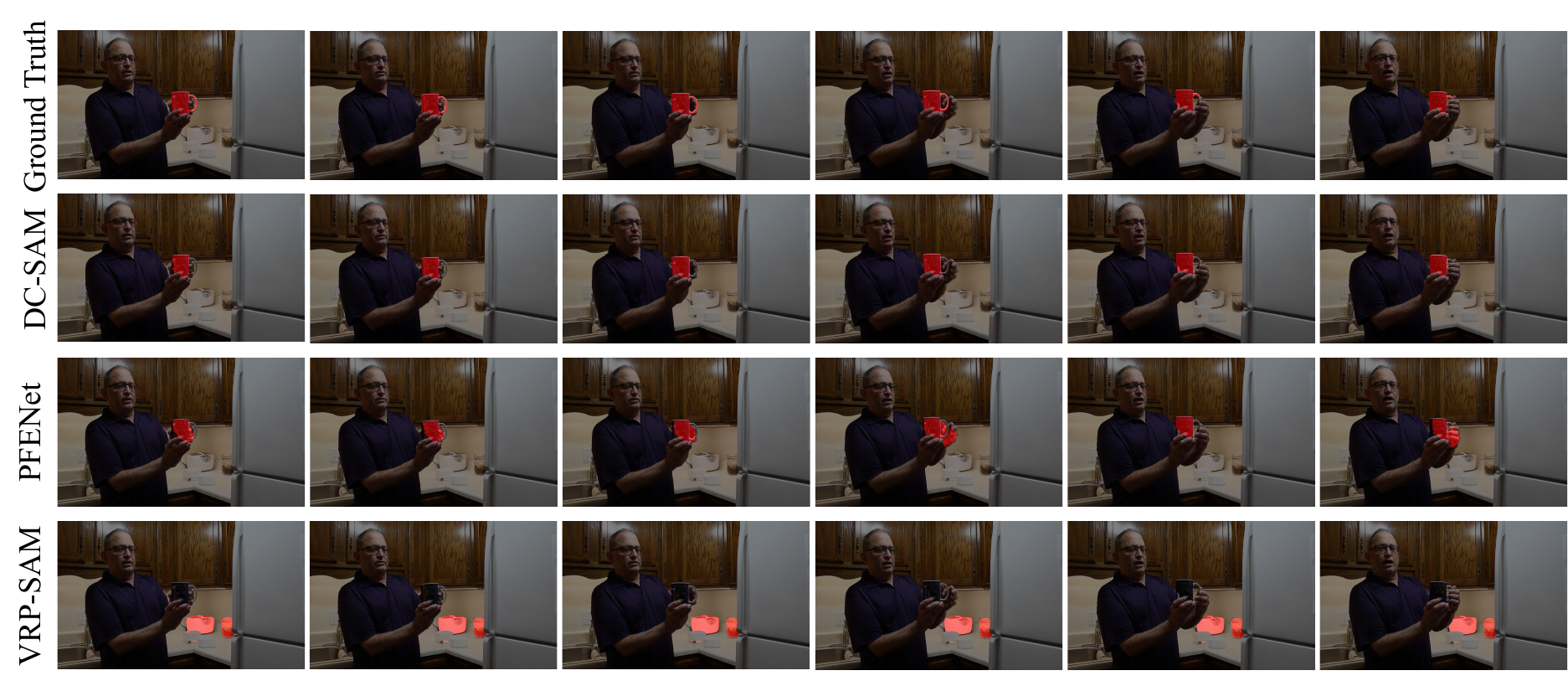}
   \end{center}
   \vspace{-3mm}
      \caption{Comparison of semantic segmentation results for the ``Cup'' category on the IC-VOS dataset.}
   \label{fig:video_vis4}
   \vspace{-3mm}
\end{figure*}

\end{document}